\documentclass[10pt]{IEEEtran}

\usepackage{fix-cm}
\usepackage{cite}
\usepackage{amsmath,amssymb,amsfonts}
\usepackage{algorithmic}
\usepackage{color}
\usepackage{textcomp}
\usepackage{hyperref}
 
\newif\ifDeepMIMOModel
\DeepMIMOModeltrue

\newif\ifSimpleNParamEq
\SimpleNParamEqtrue

\usepackage{lettrine}
\usepackage{newtxtext}
\usepackage{multirow}
\usepackage{longtable}
\usepackage[linesnumbered,ruled,vlined]{algorithm2e}
\usepackage{float}
\makeatletter
\let\oldlt\longtable
\let\endoldlt\endlongtable
\def\longtable{\@ifnextchar[\longtable@i \longtable@ii}
\def\longtable@i[#1]{\begin{figure}[t]
\onecolumn
\begin{minipage}{0.5\textwidth}
\oldlt[#1]
}
\def\longtable@ii{\begin{figure}[t]
\onecolumn
\begin{minipage}{0.5\textwidth}
\oldlt
}
\def\endlongtable{\endoldlt
\end{minipage}
\twocolumn
\end{figure}}
\makeatother
\usepackage{ifpdf}
\ifCLASSINFOpdf
  \usepackage[pdftex]{graphicx}
  \usepackage{graphicx,epstopdf}
  \graphicspath{{../pdf/}{../jpeg/}}
  \DeclareGraphicsExtensions{.pdf,.jpeg,.png,.eps}
\else
  \usepackage[dvips]{graphicx}
  \usepackage{graphicx,epstopdf}
  \graphicspath{{../eps/}}
  \DeclareGraphicsExtensions{}
\fi

\usepackage{tikz}
\usetikzlibrary{decorations.pathreplacing,calc}
\newcommand{\tikzmark}[1]{\tikz[overlay,remember picture] \node (#1) {};}

\newcommand*{\AddNote}[4]{%
    \begin{tikzpicture}[overlay, remember picture]
        \draw [decoration={brace,amplitude=0.5em},decorate,line width=.2mm,black]
            ($(#3)!(#1.north)!($(#3)-(0,1)$)$) --  
            ($(#3)!(#2.south)!($(#3)-(0,1)$)$)
                node [align=center, text width=2.5cm, pos=0.5, anchor=west] {#4};
    \end{tikzpicture}
}%

\usepackage{cite}
\usepackage{amsthm}
\usepackage{steinmetz}
\usepackage{amssymb}
\usepackage{balance}
\usepackage{eqparbox}
\usepackage{multirow}
\usepackage{bbm}
\usepackage{float}
\SetAlFnt{\small}

\usepackage{amsmath}

\usepackage{mathtools, nccmath}

\usepackage{booktabs, siunitx}

\usepackage{ragged2e} 

\usepackage[scaled]{DejaVuSansMono}
\usepackage[T1]{fontenc}

\usepackage{color}
\usepackage{relsize}
\usepackage{mathtools}

\newcommand{\bs}[1]{\boldsymbol{#1}}

\newcommand{\mb}[1]{\mathbf{#1}}

\usepackage{mathtools}

\SetKwInput{KwInput}{Input}                
\SetKwInput{KwOutput}{Output}
\SetKwInput{KwOutputr}{Output Regression}              
\SetKwInput{KwOutputc}{Output Classification}              

\newcommand{\bseq}{\begin{subequations}}
\newcommand{\eseq}{\end{subequations}}
\newcommand{\baln}{\begin{align}}
\newcommand{\ealn}{\end{align}}
\newcommand{\balnd}{\begin{aligned}}
\newcommand{\ealnd}{\end{aligned}}
\newcommand{\beq}{\begin{equation}}
\newcommand{\eeq}{\end{equation}}
\newcommand{\beqn}{\begin{eqnarray}}
\newcommand{\eeqn}{\end{eqnarray}}
\newcommand{\beqno}{\begin{eqnarray*}}
\newcommand{\eeqno}{\end{eqnarray*}}
\newcommand{\bma}{\begin{displaymath}}
\newcommand{\ema}{\end{displaymath}}
\newcommand{\bnu}{\begin{enumerate}}
\newcommand{\enu}{\end{enumerate}}
\newcommand{\bce}{\begin{center}}
\newcommand{\ece}{\end{center}}
\newcommand{\btb}{\begin{tabular}}
\newcommand{\etb}{\end{tabular}}
\newcommand{\ba}{\begin{array}}
\newcommand{\ea}{\end{array}}
\usepackage{footnote}
\makesavenoteenv{tabular}
\makesavenoteenv{table}
\makeatletter 
\newcommand\semiHuge{\@setfontsize\semiHuge{21.1}{27.38}}
\makeatother

\usepackage{subfigure}
\usepackage{float}

\makeatletter
\renewcommand{\p@subsection}{\thesection-}
\makeatother

\hypersetup{hidelinks}

\def\BibTeX{{\rm B\kern-.05em{\sc i\kern-.025em b}\kern-.08em
    T\kern-.1667em\lower.7ex\hbox{E}\kern-.125emX}}
\AtBeginDocument{\definecolor{tmlcncolor}{cmyk}{0.93,0.59,0.15,0.02}\definecolor{NavyBlue}{RGB}{0,86,125}}

\usepackage[acronym]{glossaries}
\newacronym{BS}{BS}{base station}
\newacronym{PS}{PS}{phase-shifter}
\newacronym{RL}{RL}{reinforcement learning}
\newacronym{AP}{AP}{analog precoder}
\newacronym{FC-HBF}{FC-HBF}{fully-connected HBF}
\newacronym{FSA-HBF}{FSA-HBF}{fixed subarray HBF}
\newacronym{DSA-HBF}{DSA-HBF}{dynamic subarray HBF}
\newacronym{BF}{BF}{beamforming}
\newacronym{UE}{UE}{user equipment}
\newacronym{AWGN}{AWGN}{additive white gaussian noise}
\newacronym{MIMO}{MIMO}{multiple-input multiple-output}
\newacronym{MISO}{MISO}{multiple-input single-output}
\newacronym{RF}{RF}{radio frequency}
\newacronym{RIS}{RIS}{reconfigurable intelligent surfaces}
\newacronym{IOT}{IOT}{internet-of-things}
\newacronym{CL}{CL}{convolutional layer}
\newacronym{FDD}{FDD}{frequency-division duplex}
\newacronym{TDD}{TDD}{time-division duplex}
\newacronym{CSI}{CSI}{channel state information}
\newacronym{DNN}{DNN}{deep neural network}
\newacronym{DP}{DP}{digital precoder}
\newacronym{DL}{DL}{deep learning}
\newacronym{SVD}{SVD}{singular-value decomposition}
\newacronym{CNN}{CNN}{convolution neural network}
\newacronym{FDP}{FDP}{fully digital precoder}
\newacronym{SE}{SE}{spectral efficiency}
\newacronym{OFDM}{OFDM}{orthogonal frequency division multiplexing}
\newacronym{OMP}{OMP}{orthogonal matching pursuit}
\newacronym{FL}{FL}{fully-connected layer}
\newacronym{HSHO}{HSHO}{Hybrid Structured Heuristic Optimization}
\newacronym{HBF}{HBF}{hybrid beamforming}
\newacronym{IA}{IA}{initial access}
\newacronym{mm-Wave}{mm-Wave}{millimeter wave}
\newacronym{mMIMO}{mMIMO}{massive multiple-input multiple-output}
\newacronym{SINR}{SINR}{signal-to-interference-noise ratio}
\newacronym{SNR}{SNR}{signal-to-noise ratio}
\newacronym{RSSI}{RSSI}{received signal strength indicator}
\newacronym{PZF}{PZF}{phase zero forcing}
\newacronym{PSO}{PSO}{particle swarm optimization}
\newacronym{ZF}{ZF}{zero forcing}
\newacronym{O-FDP}{O-FDP}{optimal fully digital precoder}
\newacronym{JT}{JT}{joint transmission}
\newacronym{CU}{CU}{central unit}
\newacronym{MSE}{MSE}{mean squared error}
\newacronym{CEL}{CEL}{cross entropy loss}
\newacronym{CB}{CB}{conjugate beamforming}
\newacronym{NC}{NC}{network controller}
\newacronym{CoMP}{CoMP}{coordinated multi point}
\newacronym{CF-mMIMO}{CF-mMIMO}{cell-free massive MIMO}
\newacronym{CF-HBF}{CF-HBF}{cell-free hybrid beamforming}
\newacronym{CF-BF}{CF-BF}{cell-free beamforming}
\newacronym{MLDG}{MLDG}{meta-learning domain generalization}
\newacronym{MAML}{MAML}{model agnostic meta-learning}
\newacronym{WSR}{WSR}{weighted sum rate}
\newacronym{WMMSE}{WMMSE}{weighted minimum mean squared error}
\newacronym{NN}{NN}{neural network}
\newacronym{LOS}{LOS}{line-of-sight}
\newacronym{NLOS}{NLOS}{non-line-of-sight}
\newacronym{ML}{ML}{machine learning}
\newacronym{FCL}{FCL}{fully connected layer}
\newacronym{SSL}{SSL}{semi-supervised learning}
\newacronym{MAC}{MAC}{multiply-accumulate}

\begin{document}

\markboth{}{Karkan {et al.} A Low-Complexity Plug-and-Play DL Model for mMIMO Precoding Across Sites}

\title{A Low-Complexity Plug-and-Play Deep Learning Model for Generalizable Massive MIMO Precoding}

\author{\IEEEauthorblockN{Ali~Hasanzadeh~Karkan$^{*}$, Ahmed~Ibrahim$^{\dagger}$, Jean-François~Frigon$^{*}$, and François~Leduc-Primeau$^{*}$\\
$^{*}$Department of Electrical Engineering, Polytechnique Montréal, Montréal, QC H3C 3A7, Canada\\
$^{*}$Emails: \{ali.hasanzadeh-karkan, j-f.frigon, francois.leduc-primeau\}@polymtl.ca. \\
$^{\dagger}$Ericsson Canada's R\&D,  Kanata, ON K2K 2V6, Canada, Email: ahmed.a.ibrahim@ericsson.com.
}
}
\maketitle
\IEEEpubidadjcol

\begin{abstract}
Massive multiple-input multiple-output (mMIMO) downlink precoding offers high spectral efficiency but remains challenging to deploy in practice because near-optimal algorithms such as the weighted minimum mean squared error (WMMSE) are computationally expensive, and sensitive to SNR and channel-estimation quality, while existing deep learning (DL)-based solutions often lack robustness and require retraining for each deployment site. This paper proposes a plug-and-play precoder (PaPP), a DL framework with a backbone that can be trained for either fully digital (FDP) or hybrid beamforming (HBF) precoding and reused across sites, transmit-power levels, and with varying amounts of channel estimation error, avoiding the need to train a new model from scratch at each deployment. PaPP combines a high-capacity teacher and a compact student with a self-supervised loss that balances teacher imitation and normalized sum-rate, trained using meta-learning domain-generalization and transmit-power-aware input normalization.
Numerical results on ray-tracing data from three unseen sites show that the PaPP FDP and HBF models both outperform conventional and deep learning baselines, after fine-tuning with a small set of local unlabeled samples. 
Across both architectures, PaPP achieves more than 21$\times$ reduction in modeled computation energy and maintains good performance under channel-estimation errors, making it a practical solution for energy-efficient mMIMO precoding.

\end{abstract}


\maketitle

\section{INTRODUCTION} \label{Sec:Intro}
The demand for high-capacity wireless links has intensified the adoption of \gls{mMIMO} systems, where \gls{BS} equipped with large antenna arrays serve many users simultaneously. In downlink \gls{mMIMO} systems, linear precoding plays a central role in spatially multiplexing user data streams to enhance spectral efficiency and overall system throughput. The design of such precoders is typically formulated as a sum-rate maximization problem, where the objective is to jointly optimize the beamforming vectors across all users under a transmit power constraint. However, this optimization problem is non-convex and classified as NP-hard \cite{luo2008dynamic}, making exact solutions intractable for large-scale antenna arrays. Conventional approaches rely on iterative algorithms, with the \gls{WMMSE} method \cite{shi2011iteratively} being one of the most widely adopted due to its convergence guarantees and empirical effectiveness across diverse scenarios. Despite its performance, \gls{WMMSE} incurs substantial computational cost at inference time, especially due to repeated matrix inversions whose complexity scales cubically with the number of antennas ($\mathcal{O}(N_{\sf{T}}^{3})$), rendering it impractical for real-time applications in dense or power-constrained deployments.

\Gls{DL} methods design precoding by learning a direct mapping from \gls{CSI} to beamforming weights \cite{hojatian2021unsupervised, chowdhury2021unfolding}. \Glspl{DNN} trained offline can infer those weights, slashing the run-time cost of iterative solvers. Yet this efficiency comes at the price of robustness. Models optimized for one propagation scenario may falter when the deployment site, antenna layout, \gls{SNR}, or \gls{CSI} quality differs from training. Real networks routinely face such domain shifts alongside hardware constraints and strict energy budgets, so a practical solution must pair fast inference with mechanisms that sustain performance across diverse conditions.
In this paper, we propose a learning-based precoding framework that is applicable to both \gls{FDP} and \gls{HBF} transmitter architectures. We consider three generalization goals. 
The first is site generalization. The precoder should maintain high spectral efficiency when transferred to \glspl{BS} deployed in new urban, suburban, or industrial landscapes~\cite{liu2025generalizing}. 
The second is \gls{SNR} independence. The precoder must adapt gracefully across a broad \gls{SNR} range, for instance, resulting from variations in the base station transmit power, without re-training. 
Third, real systems rely on estimated channels that are corrupted by noise, quantization, and pilot contamination. The design should remain effective under such \gls{CSI} impairments~\cite{lavi2023learn}.
Finally, we seek to reduce computational energy compared to conventional methods.

To study these challenges under realistic propagation conditions, we leverage a large-scale ray-tracing data set generated from detailed three-dimensional maps of Montreal. This dataset includes several base-station sites, ranging from industrial campuses to dense downtown streets, in order to adequately benchmark the generalization capability of learned precoding methods.

\subsection{RELATED WORK}\label{sec:related}
Precoding for \gls{mMIMO} systems has evolved from classical optimization to modern deep learning techniques. Traditional linear precoding methods like eigen-beamforming~\cite{vu2007mimo} and zero-forcing variations~\cite{liu2019near,zhang2021joint} laid the groundwork but are often limited by their reliance on perfect \gls{CSI} and by their computational complexity. 
Non-linear methods~\cite{zou2023capacity,zeng2019novel,flores2021tomlinson,krishnamoorthy2022downlink,jiang2022deep} can theoretically approach channel capacity but are often impractical due to high complexity and sensitivity to channel estimation errors. To balance performance and hardware costs, hybrid analog-digital architectures have been explored, though early designs were iterative and hardware-specific~\cite{kabalci2022optimal}. While some works have explored specialized contexts like cache-aided precoding~\cite{wei2019cache}, satellite links~\cite{ha2022joint}, and randomized iterations~\cite{wang2022statistical}, the core challenges of scalability and robustness remain.

Deep learning has emerged as a powerful alternative to slash the computational burden of conventional precoders~\cite{shi2011iteratively, shi2021deep}. Early work focused on learning a direct mapping from \gls{CSI} to beamforming weights, often by imitating classical solvers like \gls{WMMSE}~\cite{shi2021deep}. However, these models often lack robustness, failing to generalize to deployment conditions not seen during training. This has spurred research into more robust and adaptable solutions~\cite{hojatian2021unsupervised,chowdhury2021unfolding}. Hybrid analog–digital precoders further complicate matters by introducing hardware constraints such as finite-resolution phase shifters and limited \gls{RF} chains~\cite{hojatian2024learning}.

A significant body of work has focused on enhancing the robustness of learned precoders. Meta-learning approaches, such as \gls{MAML}~\cite{lyu2023downlink} and \gls{MLDG}~\cite{karkan2024sage}, have been utilized to train models that can rapidly adapt to new environments. Adversarial training has also been employed to create models resilient to worst-case \gls{CSI} perturbations~\cite{liu2025adversarial}. For hybrid architectures, some research has focused on jointly optimizing analog and digital precoders under hardware constraints 
and compressing \gls{CSI} feedback~\cite{shruthi2025hybrid, cui2024lightweight}.

Recent efforts have also explored novel neural network architectures. Graph neural networks have been used to model the relationship between antennas and users, enabling hardware-aware precoding~\cite{feys2024toward}. Transformers have been applied to capture global correlations across antennas and users in fully digital systems~\cite{jiang2023transformer}, while other works have focused on reducing pilot overhead in \gls{FDD} systems~\cite{song2023deep}. Despite these advances, many learned designs are evaluated under idealized conditions, assuming perfect \gls{CSI}~\cite{akrout2023domain}.

\subsection{CONTRIBUTIONS}
This paper proposes a low-complexity ``plug-and-play precoder'' (PaPP) that simultaneously addresses site generalization, robustness, efficiency, and architectural flexibility\footnote{A preliminary version of this work was presented at the ICMLCN 2025 conference \cite{hasanzadeh2025low}.}. Specifically, PaPP maintains high spectral efficiency across \emph{unseen} deployment sites in diverse urban, suburban, and industrial scenarios, while remaining \emph{SNR-independent} and resilient to noisy and imperfect \gls{CSI}. It further achieves substantial computational savings compared to conventional iterative approaches such as \gls{WMMSE}, and its design accommodates both \gls{FDP} and \gls{HBF} transmitter architectures, enabling deployment under different hardware constraints. These capabilities stem from a teacher–student framework combined with self-supervised learning, supported by \gls{MLDG} for improved generalization. Collectively, the results demonstrate that PaPP provides a unified and practical solution to the multifaceted challenges of next-generation \gls{mMIMO} systems.

\subsection{PAPER ORGANIZATION AND NOTATION}
The remainder of this paper is organized as follows. Section~\ref{Sec:Problem_Formulation} details the system model and problem formulation. Section~\ref{Sec:Proposed} introduces our proposed PaPP model, including its architecture and training methodology. Section~\ref{Sec:Complexity} provides a comparative analysis of computational complexity against baseline methods. Section~\ref{Sec:Simulation} presents the experimental setup and discusses the performance of PaPP under various conditions. Finally, Section~\ref{Sec:conclusion} concludes the paper and outlines future research directions.

Throughout this paper, we use the following notation. Boldface lowercase and uppercase letters denote vectors and matrices, respectively (e.g., $\mathbf{w}$ and $\mathbf{W}$). The conjugate transpose of a matrix $\mathbf{A}$ is denoted by $\mathbf{A}^{\dagger}$. The trace of a square matrix is denoted by $\mathrm{Tr}(\cdot)$. The expectation operator is denoted by $\mathbb{E}[\cdot]$. The set of complex numbers is denoted by $\mathbb{C}$. The distribution of a circularly symmetric complex Gaussian random vector with mean $\boldsymbol{\mu}$ and covariance matrix $\mathbf{\Sigma}$ is denoted by $\mathcal{CN}(\boldsymbol{\mu}, \mathbf{\Sigma})$. $\mathbf{I}$ denotes an identity matrix of appropriate dimensions.

\section{PROBLEM FORMULATION} \label{Sec:Problem_Formulation}
\subsection{SYSTEM MODEL}
In this work, we focus on a \gls{TDD} multi-user \gls{mMIMO} system. In \gls{TDD}, the \gls{BS} can infer the downlink channel from uplink pilots without incurring additional feedback overhead. The channel matrix $\mathbf{H}\in\mathbb{C}^{N_{\sf U}\times N_{\sf T}}$ estimated during the uplink slot is reused to design the subsequent downlink precoder. The \gls{BS} is equipped with $N_{\sf T}$ antennas and simultaneously serves $N_{\sf U}$ single-antenna users. Typically $N_{\sf U} \ll N_{\sf T}$ so that a large spatial aperture is available for beamforming and interference suppression. By steering $N_{\sf U}$ independent data streams through appropriately chosen beamforming vectors, the array turns spatial degrees of freedom into multiplexing gain, boosting spectral efficiency and delivering near-orthogonal links to all users within the same time–frequency resource block.

\subsubsection{FULLY DIGITAL PRECODER}\label{subsec:fdp}
In the fully digital architecture, every transmit antenna is driven by its own \gls{RF} chain, permitting arbitrary complex weighting of each antenna signal. Under this model, the received baseband signal at the $k^{\text{th}}$ user is  
\begin{equation}\label{eq:signal_recived_braced}
\mathbf{y}_k \;=\;
    \underbrace{\mathbf{h}_{k}^{\dagger}\mathbf{w}_{k}x_k}_{\text{desired signal}}
    \;+\;
    \underbrace{\mathbf{h}_{k}^{\dagger}\!\!\sum_{j\neq k}\mathbf{w}_{j}x_j
                \;+\;\bs{\eta}_k}_{\text{multi‐user interference + noise}} ,
\end{equation}
where $\mathbf{h}_{k}\!\in\!\mathbb{C}^{N_{\sf{T}}\times 1}$ denotes the downlink channel from the \gls{BS} to user~$k$,  
$x_k$ is the user’s data symbol with $\mathbb{E}[\mb{x}\mb{x}^{\dagger}] = \frac{1}{N_{\sf{U}}}\mathbbm{1}_{N_{\sf{U}}}$, and $\bs{\eta}_k\!\sim\!\mathcal{CN}(0,\sigma^{2})$.  
Collecting all user beamformers gives the \gls{FDP} matrix  
$\mathbf{W}= [\mathbf{w}_{1},\ldots,\mathbf{w}_{N_{\sf U}}]\!\in\!\mathbb{C}^{N_{\sf T}\times N_{\sf U}}$.  
The instantaneous \gls{SINR} of user~$k$ is therefore  
\begin{equation}\label{eq:sinr}
    \text{SINR}(\mathbf{w}_{k}) 
      = \frac{\lvert\mathbf{h}_{k}^{\dagger}\mathbf{w}_{k}\rvert^{2}}
             {\displaystyle\sum_{j\neq k}\lvert\mathbf{h}_{k}^{\dagger}\mathbf{w}_{j}\rvert^{2}+\sigma^{2}}\,,
\end{equation}
and the corresponding network capacity (sum-rate) is  
\begin{equation}\label{eq:sum-rate-fdp}
    R(\mathbf{W}) = \sum_{\forall k}\log_{2}\!\bigl(1+\text{SINR}(\mathbf{w}_{k})\bigr)\,.
\end{equation}
Maximising~\eqref{eq:sum-rate-fdp} subject to a total power budget $P_{\sf Tx}$ gives the \gls{FDP} problem
\begin{equation}\label{eq:maximization_fdp}
\begin{aligned}
   &\underset{\mathbf{W}}{\max}~ R(\mathbf{W}),\\
   \text{s.t.} &\;\sum_{\forall k}\mathbf{w}_{k}^{\dagger}\mathbf{w}_{k}\le P_{\sf Tx}\,.
\end{aligned}
\end{equation}

\subsubsection{HYBRID ANALOG–DIGITAL PRECODER}\label{subsec:hbf}
Extending the \gls{FDP} setting, the \gls{BS} still serves $N_{\sf U}$ single-antenna users through $N_{\sf T}$ antennas, but now employs only $N_{\sf RF}\!\ll\!N_{\sf T}$ RF chains. A digital precoder first forms $N_{\sf RF}$ streams, which a constant-modulus phase-shifter network $\mb A\!\in\!\mathbb{C}^{N_{\sf T}\times N_{\sf RF}}$ spreads across the full array, corresponding to a fully connected \gls{HBF} architecture where each RF chain drives all antennas through phase shifters. Setting $\mb A=\mb I_{N_{\sf T}}$ and $N_{\sf RF}=N_{\sf T}$ recovers the \gls{FDP}. With the downlink channel vector $\mb h_u\!\in\!\mathbb{C}^{N_{\sf T}\times1}$, digital beamformers $\mb W=[\mb w_1,\ldots,\mb w_{N_{\sf U}}]\!\in\!\mathbb{C}^{N_{\sf RF}\times N_{\sf U}}$, user symbol $x_u$ (from $\mb x=[x_1,\ldots,x_{N_{\sf U}}]^\top$, $\mathbb{E}[\mb x\mb x^\dagger]=\frac1{N_{\sf U}}\mathbbm 1_{N_{\sf U}}$), and AWGN $\bs\eta_u\!\sim\!\mathcal{CN}(0,\sigma^2)$, the received signal is
\begin{equation}\label{recv_sig_braced}
\mb y_u
\;=\;\underbrace{\mb h_u^{\dagger}\mb A\mb w_u x_u}_{\text{desired signal}}
\;+\;\underbrace{\mb h_u^{\dagger}\mb A\!\!\sum_{j\neq u}\mb w_j x_j+\bs\eta_u}_{\text{interference+noise}} \,,
\end{equation}
where each entry of $\mb A$ satisfies $|[\mb A]_{n,m}|=1$ with phase $e^{j\varphi_{n,m}}$. Hence, hybrid beamforming retains the linear signal model while substantially reducing \gls{RF}-chain count and power consumption. This saving becomes even more critical at millimetre-wave and sub-THz frequencies.

In \eqref{recv_sig_braced}, the first bracketed term represents the desired signal component intended for user $u$, while the second bracketed term aggregates undesired signals.
\Gls{HBF} seeks to jointly design both the digital precoder $\mb{W}$ and the analog precoder $\mb{A}$ to maximize signal quality and suppress interference, all while adhering to hardware constraints and minimizing power consumption. 
The SINR for user~$u$ is given by 
\begin{equation}\label{eq:SINR_HBF_re}
    \text{SINR}(\mb{A},\mb{w}_{u})
      = \frac{\lvert\mb{h}_{u}^{\dagger}\mb{A}\mb{w}_{u}\rvert^{2}}
             {\displaystyle\sum_{j\neq u}\lvert\mb{h}_{u}^{\dagger}\mb{A}\mb{w}_{j}\rvert^{2}+\sigma^{2}},
\end{equation}
and the aggregate capacity is  
\begin{equation}\label{eq:sumRate_HBF_re}
    R(\mb{A},\mb{W}) = \sum_{\forall u}\log_{2}\!\bigl(1+\text{SINR}(\mb{A},\mb{w}_{u})\bigr).
\end{equation}
Jointly designing the analog and digital components involves  
\begin{subequations}\label{eq:hbf_opt_final}
\begin{align}
  &\underset{\mb{A},\mb{W}}{\max}\; R(\mb{A},\mb{W})\\
  \text{s.t.}&\;
      \sum_{\forall u}\mb{w}_{u}^{\dagger}\mb{A}^{\dagger}\mb{A}\mb{w}_{u}\le P_{\sf Tx}\,.
\end{align}
\end{subequations}

\subsection{WMMSE ALGORITHM}
The optimization problem in~\eqref{eq:maximization_fdp} is non-convex and NP-hard. A widely adopted heuristic is the iterative \gls{WMMSE} algorithm~\cite{shi2011iteratively}, which reformulates sum-rate maximization as minimization of the total \gls{MSE} under the independence of $x_k$ and $\bs{\eta}_k$. This reformulation introduces two auxiliary variables per user: the linear receiver gain $u_k$ and a positive weight $v_k$. Together they define the per-user \gls{MSE} covariance, and the algorithm alternates among three convex sub-problems that update the receivers ${u_k}$, the weights ${v_k}$, and the precoder~$\mathbf W$.

Starting from an initial precoder $\mathbf W^{(0)}$ that satisfies the power constraint in~\eqref{eq:maximization_fdp}, the updates at iteration $i$ are
\begin{align}
    v^{(\text{i})}_{k} = \frac{ \sum\limits_{\substack{j =1}}^{N_{\sf{U}}} \big|\mb{h}^{\dagger}_{k} \mb{w}^{(\text{i}-1)}_{j} \big|^2 + \sigma^2}{\sum\limits_{\substack{j \neq k}}^{N_{\sf{U}}} \big|\mb{h}^{\dagger}_{k} \mb{w}^{(\text{i}-1)}_{j} \big|^2 + \sigma^2},
\end{align}

\begin{align}
    u^{(\text{i})}_{k} = \frac{ \mb{h}^{\dagger}_{k} \mb{w}^{(\text{i}-1)}_{j} }{\sum\limits_{\substack{j=1}}^{N_{\sf{U}}} \big|\mb{h}^{\dagger}_{k} \mb{w}^{(\text{i}-1)}_{j} \big|^2 + \sigma^2},
\end{align}
\begin{align}\label{eq:construct_beamforming_matrix}
    \mb{w}^{(\text{i}+1)}_{k} =  u^{(\text{i})}_{k} v^{(\text{i})}_{k}\mb{h}_{k}\left(\sum^{N_{\sf{U}}}_{j=1}   v^{(\text{i})}_{j} |u^{(\text{i})}_{j}|^2 \mb{h}_{j}\mb{h}_{j}^{\dagger} + \mu \mathbf{I}\right)^{-1},
\end{align}
where $\mu\!\ge\!0$ is the Lagrange multiplier guaranteeing the transmit-power constraint. The iterations continue until a chosen convergence criterion (e.g., change in sum-rate) is met.

\section{THE PaPP MODEL} \label{Sec:Proposed}
We present a \gls{DL}-based precoding framework that has the ability to generalize to previously unseen operating conditions. We consider generalization over three aspects of the deployment: the base-station site, the base-station transmit power, and the amount of estimation error on the CSI input.
The proposed strategy involves first training a general ``backbone'' model. The model deployed to a particular base station is then obtained by keeping only a part of the backbone model. Crucially, this training method enables implicit adaptation to varying channel qualities without explicit SNR inputs, allowing the model to approximate WMMSE performance with significantly reduced computational overhead. Optionally, a deployed model can also be fine-tuned using local data for improved performance.

We first describe the backbone DNN architecture, and then present various strategies that enable it to reach excellent generalization performance. Finally, we discuss the deployment and optional fine-tuning.

\subsection{NEURAL NETWORK ARCHITECTURE}
\begin{figure}[t]
    \centering
    \includegraphics[width=\linewidth]{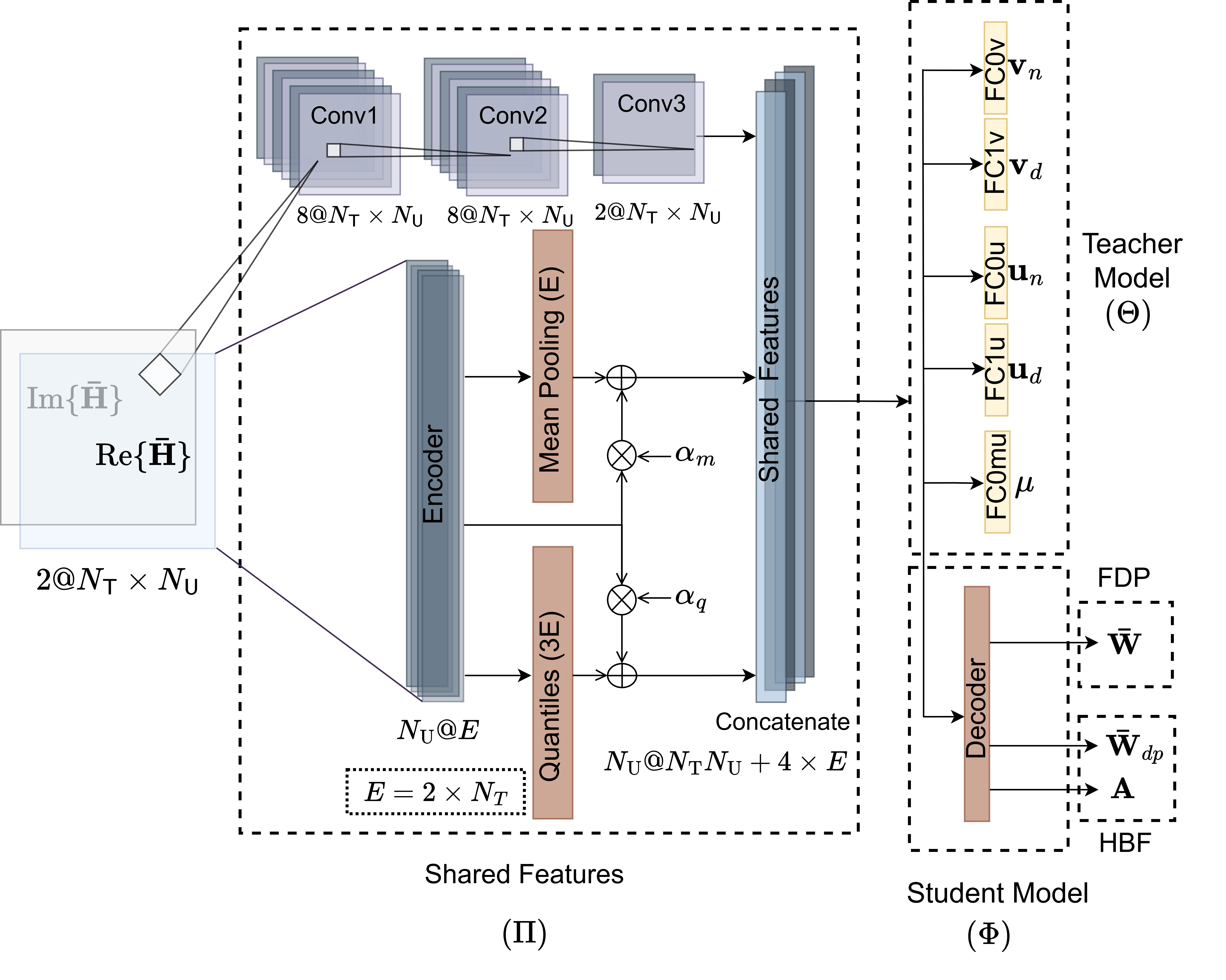}
    \caption{Proposed ``backbone'' \gls{DNN} architecture. Only one student head (FDP or HBF) is activated at a time.}
    \label{fig:arch}
\end{figure}

The primary objective of the proposed PaPP is to obtain a computationally efficient \gls{DNN} that replaces the expensive matrix inversion in \eqref{eq:construct_beamforming_matrix}. However, our experiments indicate that naïve self-supervised training lags significantly behind \gls{WMMSE} performance on unseen channels. To address this, we employ a teacher-student architecture coupled with \gls{MLDG}. A high-capacity teacher learns an accurate WMMSE surrogate, while a lightweight student distills that knowledge without performing matrix inversions.

The model takes instantaneous channel estimates, obtained from uplink pilots in \gls{TDD}, as input and directly outputs the final downlink precoding. In the \gls{FDP} case, the framework produces the digital precoding matrix, while in the \gls{HBF} case, it generates the analog and digital components separately. 

The proposed backbone DNN architecture is illustrated in Figure~\ref{fig:arch}. Its input is a scaled version of the complex-valued channel state matrix, denoted $\bar{\mathbf{H}}$, which will be described in Section~\ref{sec:snr_scaling}. We extract features using two parallel branches to capture both spatial and user-specific contexts.

First, to extract global features, we process the channel estimates through a three-layer convolutional network with output channels (8, 8, 2). Each convolutional layer applies small receptive fields over the spatial and antenna dimensions. Stacking these layers allows deep feature hierarchies to emerge, making the extractor effective at modeling both local channel variations and complex inter-user interference patterns.

Second, to extract per-user features, we 
concatenate the real and imaginary parts of $\mathbf{h}_k$ for each user $k$. The resulting vector of length $E\triangleq 2N_T$ is fed into a five-layer MLP encoder. The outputs of these $N_U$ encoders are then pooled across users using four permutation-invariant statistics (the feature-wise mean and three fixed quantiles) to form a global context vector of length $4E$. This summary captures typical link quality and user heterogeneity. 

To merge this context back into the user embeddings, the context vector is broadcast to match the user dimension and combined with the user embeddings via learnable scalar gates ($\alpha_m$ for the mean, $\alpha_q$ for the quantiles). 
This combination of the context with the initial user embeddings can be seen as a residual connection, which eases optimization by preserving identity mappings \cite{he2016deep}, while gating mechanisms regulate the integration of global summary information. For this merge, the user embeddings are first duplicated 4 times to match the length of the statistics vector.

Finally, the two feature extraction branches are combined into shared feature vectors. For this, the output of the CNN extractor must be replicated $N_U$ times to match the dimension of the user embeddings.
Note that we found empirically that the redundant representations that are introduced within the feature extraction were necessary to improve the training outcomes.

From the shared features, the \textit{teacher} branch predicts the WMMSE auxiliary variables $\mathbf{v}$, $\mathbf{u}$, and $\mu$ through linear heads. To decouple the teacher output from the noise power, the DNN has separate outputs for the numerators and denominators of $\mathbf{v}$ and $\mathbf{u}$, which are then used to form
\begin{align}
    v_k &= \frac{v_{n,k} + \sigma^2}{v_{d,k} + \sigma^2}\,, \label{eq:interWMMSEv}\\[6pt]
    u_k &= \frac{u_{n,k}}{u_{d,k} + \sigma^2}\,. \label{eq:interWMMSEu}
\end{align}

The same shared features feed the lightweight \textit{student} branch, which distills the teacher’s outputs into the final precoder used at inference.
The student model utilizes distinct output heads to adapt to the specific transmitter architecture.
In the \gls{FDP} configuration, the model outputs $\mathbf{\bar{W}}\!\in\!\mathbb{C}^{N_{\sf T}\times N_{\sf U}}$, representing the user-to-antenna mapping. 
In the \gls{HBF} configuration, the model outputs the digital precoder $\mathbf{\bar{W}}_{dp}\!\in\!\mathbb{C}^{N_{\sf RF}\times N_{\sf U}}$ and the analog phase shifts $\mathbf{A}\!\in\!\mathbb{C}^{N_{\sf T}\times N_{\sf RF}}$.
The predicted precoder is normalized to satisfy a unit transmit-power constraint, such that 
\begin{equation}
\operatorname{Tr}\!\left(\bar{\mathbf{W}}\bar{\mathbf{W}}^{\mathsf H}\right)=1 \,.
  \label{eq:unit-power-normalization}
\end{equation}

\subsection{DOWNLINK POWER ADAPTATION}\label{sec:snr_scaling}
In a practical BS, the transmit power $P_{\sf{Tx}}$ can be varied for coverage and link adaptation. Since the choice of $P_{\sf{Tx}}$ affects the precoding, we must ensure that the DNN precoder can adapt dynamically to changes in $P_{\sf{Tx}}$.
We propose to achieve this by simply scaling the channel input provided to the DNN, denoted as $\bar{\mathbf{H}}$. This input is computed as
\begin{equation}\label{eq:normalized}
    \bar{\mathbf H} \;\triangleq\; \frac{\sqrt{P_{\sf{Tx}}}}{\sigma}\,\mathbf H\,,
\end{equation}
where \(\mathbf{H}=[\mathbf{h}_1,\ldots,\mathbf{h}_{N_U}]^{\top}\).
Then, to adjust the transmit power, the actual precoder used for transmission can simply be obtained from the normalized precoder $\bar{\mathbf{W}}$ generated by the DNN as
\begin{equation}\label{eq:precodernormalized}
    \mathbf W=\sqrt{P_{\sf{Tx}}}\,\bar{\mathbf W}\,.  
\end{equation}
Therefore, from the perspective of the DNN, the SINR of any user can be computed based on $\bar{\mathbf{H}}$. We can easily verify that the SINR in \eqref{eq:sinr} is unchanged if we replace $\mathbf{H}$ by $\bar{\mathbf H}$, $\mathbf{W}$ by $\bar{\mathbf{W}}$, and $\sigma$ by $1$.
This approach enables the DNN to adapt to the chosen transmit power and to the actual noise power without adding any inputs to the DNN model.

\subsection{TRAINING LOSS}
The training pipeline combines three ingredients: a teacher–student scheme for knowledge transfer, self-supervised learning to optimize the model without labels, and meta-learning to improve generalization across sites and different \gls{SNR} conditions.

For the teacher-student training \cite{hinton2015distilling}, we leverage a high-performance teacher with parameters \(\Theta\), a shared feature extractor with parameters \(\Pi\), and an efficient student with parameters \(\Phi\). The teacher produces a precoder \(\bar{\mb W}_T\) by executing a single \gls{WMMSE} iteration seeded with intermediate variables predicted by the teacher \gls{DNN}. Re-using \eqref{eq:construct_beamforming_matrix}, we set
\[
\bar{\mb W}_T \;=\; \big[\bar{\mb w}_1^{(1)},\,\dots,\,\bar{\mb w}_{N_{\sf U}}^{(1)}\big],
\]
where \(u_k^{(0)},\,v_k^{(0)},\,\mu\) are obtained from the teacher’s output heads and using \eqref{eq:interWMMSEv} and \eqref{eq:interWMMSEu}.
The student consumes the shared features and directly predicts the precoding matrix for the selected architecture (\gls{FDP} or \gls{HBF}).

The teacher is trained in a self-supervised manner, using sum-rate maximization as the objective. This eliminates the need for labeled targets and aligns learning directly with the system metric. 
To prepare for generalization training (see Section~\ref{sec:generalization}), we normalize the sum-rate of the generated precoder by its \gls{WMMSE} baseline computed on the same channel realization:
\begin{align}
    \mathcal{L}_{T}
    \;=\;
    -\,\frac{R(\bar{\mb{W}}_{T})}{R_{\text{WMMSE}}}\,,
\end{align}
where \(R_{\text{WMMSE}}\) denotes the sum-rate achieved by \gls{WMMSE}. This normalization mitigates scale disparities in achievable rates across training domains, thereby improving convergence.

For the student, we combine imitation of the teacher with direct sum-rate improvement. When the chosen architecture is \gls{HBF}, the composite precoder is formed as
\begin{align}
    \bar{\mb{W}} \;=\; \mb{A}\, \bar{\mb{W}}_{dp}.
\end{align}
The student loss is
\begin{align}\label{eq:studentlossfunction}
    \mathcal{L}_{S}
    \;=\;
    \mathcal{L}_{\text{MSE}}(\bar{\mb{W}}_{T},\bar{\mb{W}})
    \;-\;
    \lambda\,\frac{R(\bar{\mb{W}})}{R_{\text{WMMSE}}},
\end{align}
with the imitation term
\begin{align}
    \mathcal{L}_{\text{MSE}}(\bar{\mb{W}}_{T},\mb{W})
    \;=\;
    \frac{1}{N_{\sf T}N_{\sf U}}
    \sum_{i=1}^{N_{\sf T}}\sum_{j=1}^{N_{\sf U}}
    \big(\bar{\mb{W}}_{T,ij}-\bar{\mb{W}}_{ij}\big)^{2}\,.
\end{align}
We adapt the balance between imitation and rate optimization during training via a reliability gate.
We define two hyperparameters, $0\leq\lambda_0\leq\lambda_1$, to assign different weights to the student self-supervision. We set $\lambda = \lambda_1$ whenever, over the last $P$ epochs, the student's imitation loss on the validation set improves by less than $\tau$ relative to the running best (indicating a plateau), to encourage overall improvement. Otherwise, training emphasizes stable imitation by using $\lambda=\lambda_0$.

\subsection{CHANNEL ESTIMATION ERROR}
In a real deployment, as opposed to a simulation, CSI is obtained through an estimation process that is subject to an unavoidable error. We model the channel estimation error using a simple error model for an unbiased estimator as \cite{6375940}
\begin{equation}\label{eq:estimated_channel}
\hat{\mathbf{H}} = \sqrt{1-\beta^{2}}\,\mathbf{H} + \beta\,\boldsymbol{\epsilon}\,,
\end{equation}
where \(\mathbf{H}\) denotes the true channel matrix, \(\beta\in[0,1]\) controls the amount of noise on the estimate, and \(\boldsymbol{\epsilon}\sim\mathcal{N}(\mathbf{0},\sigma_{\epsilon}^{2})\). 
We select $\sigma_{\epsilon}^2 = \sigma_{H}^2$, where $\sigma_{H}^2$ denotes the variance of the true channel $\mathbf{H}$, so that the total power of the estimated channel remains equal to that of the true channel regardless of the error weighting parameter $\beta$.
When considering a channel estimation error, the CSI input of the DNN model is obtained as usual by using \eqref{eq:normalized}, but replacing $\mathbf{H}$ with $\hat{\mathbf{H}}$.

Noisy CSI data is used in two different ways. First, when training the backbone model, we consider that we have access to perfect CSI, since the backbone may be trained using a simulated dataset. However, we may wish nonetheless to simulate a noisy CSI input as part of the training. In this case, we control the value of $\beta$. Therefore, we define a set $\mathcal{B}_\textsc{BB}$ containing $\beta$ values used during backbone training, which acts as a training hyperparameter.
Second, at deployment, we consider that the deployment environment sets the value of $\beta$. 
Therefore, any fine-tuning dataset collected locally is affected by an estimation error, which has implications for the self-supervised fine-tuning of the model.

\subsection{TRAINING FOR GENERALIZATION}\label{sec:generalization}
The core of the proposed framework is the deployment-agnostic precoding backbone. This model is trained for a single transmitter architecture at a time, either \gls{FDP} or \gls{HBF}, by activating only the corresponding output heads and loss functions while disabling the others.
Regardless of the specific architecture,  the pervasive goal is to ensure the model generalizes effectively across a wide range of channel conditions. 

To achieve this robust generalization, the backbone is exposed during training to diverse physical sites with varying propagation statistics, a broad range of transmit power levels ($P_{\sf{Tx}}$), and varying degrees of channel estimation error.
For transmit power adaptation, we encode all CSI input using \eqref{eq:normalized}, and training samples span a set of \(P_{\sf{Tx}}\)  values uniformly spread over the desired operating conditions. Similarly, to account for the imperfect channel estimation, the training data includes a discrete set of estimation error levels $\mathcal{B}_\textsc{BB}$.

To enhance adaptability, we adopt \gls{MLDG} as in \cite{karkan2024sage}. 
Unlike standard mini-batch learning, \gls{MLDG} inserts an explicit meta-train/meta-generalization split at every iteration to mimic the distribution shift faced at deployment.
We formally define the source domains ($\bs{\mathcal{D}}$) based on these variations: the number of training sites (\(N_s\)), the number of transmit power values (\(N_p\)), the propagation condition (either \gls{LOS} or \gls{NLOS}), and the number of estimation error levels ($N_\beta=|\mathcal{B}_\textsc{BB}|$). Accordingly, the total number of distinct domains in the training set is given by
\begin{equation}
\bigl|\bs{\mathcal{D}}\bigr| \;=\; 2\,N_s\,N_p\,N_\beta.
\label{eq:trainset_size}
\end{equation}

The overall PaPP training procedure is summarized in Alg.~\ref{Alg1}.
We randomly form two disjoint subsets from the source domain $\bs{\mathcal{D}}$: a training subset $\bs{\mathcal{D}}^{\text{train}}$ used to update the model, and a generalization subset $\bs{\mathcal{D}}^{\text{gen}}$ used immediately afterward to assess how well that update transfers to different operating conditions. 

Both the teacher and the student follow the same meta-train/generalization pattern. Lines~\ref{line:3} and~\ref{line:4} for the teacher, and lines~\ref{line:10} and~\ref{line:11} for the student, perform the inner (meta-train) update on the training split \(\bs{\mathcal D}^{\text{train}}\). The teacher loss updates $\{\Pi, \Theta\}$, whereas the student loss updates only $\Phi$.
Note that during the offline training phase, where access to perfect CSI is assumed, these loss functions are evaluated using $\mathbf{H}$ to ensure the model learns to effectively denoise the input. 
Lines~\ref{line:5}--\ref{line:7} for the teacher, and lines~\ref{line:12} and~\ref{line:13} for the student, implement the outer (meta-generalization) step of \gls{MLDG}. After evaluating the post–inner-step model on the held-out split \(\bs{\mathcal D}^{\text{gen}}\), a meta-update is performed by combining gradients from \(\bs{\mathcal D}^{\text{train}}\) and \(\bs{\mathcal D}^{\text{gen}}\). Including both terms prevents overfitting to the current training slice (preserving in-domain accuracy) while steering updates toward directions that also reduce loss on unseen channel conditions (improving out-of-domain robustness). In practice, this yields stable convergence: \(\Phi\) improves on the current batch without degrading generalization. 

\begin{algorithm}[t]
  \caption{PaPP Backbone Training}
  \label{Alg1}
  \SetInd{0.1em}{.4em} 
  \SetKwFor{Whilee}{while}{do \hfill \CommentSty{//PaPP Backbone Training}}{end}
  \SetKwProg{ProcT}{Procedure}{:\hfill \CommentSty{//Teacher}}{end}
  \SetKwProg{ProcS}{Procedure}{:\hfill \CommentSty{//Student}}{end}
  \SetKwInOut{Init}{Initialize}
  \KwIn{Different sites channel models $\bs{\mathcal{D}}$ (source domain)}
  \Init{Parameters $\Theta$ (Teacher), $\Phi$ (Student), $\Pi$ (Features)\\ 
  Learning rates $\alpha_{T}, \alpha_{S}, \alpha_{F}, \beta_{T}, \beta_{S}, \beta_{F}, \epsilon_{T}, \epsilon_{S}, \epsilon_{F}$} \ProcT{\textsc{ttrain}$(E,\bs{\mathcal{D}}^{\text{train}},\bs{\mathcal{D}}^{\text{gen}})$}{
  \For{$e\leftarrow 1$ \KwTo $E$}{
    $\delta_{T} \leftarrow  \frac{1}{{|\bs{\mathcal{D}}^{\text{train}}|}}\sum \limits_{\substack{\hat{D} \in \bs{\mathcal{D}}^{\textnormal{train}}}}\nabla_{\Theta,\Pi}\mathcal{L}_{T}(\hat{D};\Theta,\Pi)$\label{line:3}\tikzmark{top} 
    ~~\tikzmark{right}
    
    $\Theta^{\prime} = \Theta - \alpha_{T} \delta_{T}$,~
    $\Pi^{\prime} = \Pi - \alpha_{F} \delta_{T}$\label{line:4}\tikzmark{bottom}
    \AddNote{top}{bottom}{right}{~\hspace{-20pt}Meta-Training}\\
    
    $\delta_{T}^{\prime} \leftarrow \frac{1}{{|\bs{\mathcal{D}}^{\text{gen}}|}} \sum\limits_{\substack{\hat{D} \in \bs{\mathcal{D}}^{\textnormal{gen}}}} \nabla_{\Theta^{\prime},\Pi^{\prime}}\mathcal{L}_{T}(\hat{D};\Theta^{\prime},\Pi^{\prime})$\label{line:5}\tikzmark{top}

    $\Theta \leftarrow \Theta -  \epsilon_{T}(\delta_{T} + \beta_{T} \delta_{T}^{\prime})$\label{line:6}\\
    $\Pi \leftarrow \Pi -  \epsilon_{F}(\delta_{T} + \beta_{F} \delta_{T}^{\prime})$\label{line:7}\tikzmark{bottom}
    \AddNote{top}{bottom}{right}{~\hspace{-5pt}Meta-Generalizing}\\
    }
  }
  \ProcS{\textsc{strain}$(E,\bs{\mathcal{D}}^{\text{train}},\bs{\mathcal{D}}^{\text{gen}})$}{
  \For{$e\leftarrow 1$ \KwTo $E$}{
    $\delta_{S} \leftarrow \frac{1}{{|\bs{\mathcal{D}}^{\text{train}}|}} \sum \limits_{\substack{\hat{D} \in \bs{\mathcal{D}}^{\textnormal{train}}}} \nabla_{\Phi}\mathcal{L}_{S}(\hat{D}, \mb{W}_{T};\Phi)$\label{line:10}\tikzmark{top}\\
    $\Phi^{\prime} = \Phi - \alpha_{S} \delta_{S}$ \label{line:11}\tikzmark{bottom}
    \AddNote{top}{bottom}{right}{~\hspace{-20pt}Meta-Training}\\

    $\delta_{S}^{\prime} \leftarrow \frac{1}{{|\bs{\mathcal{D}}^{\text{gen}}|}}\sum\limits_{\substack{\hat{D} \in \bs{\mathcal{D}}^{\textnormal{gen}}}} \nabla_{\Phi^{\prime}}\mathcal{L}_{S}(\hat{D}, \mb{W}^{\prime}_{T};\Phi^{\prime})$\label{line:12}\tikzmark{top}\\
    $\Phi \leftarrow \Phi -  \epsilon_{S}(\delta_{S} + \beta_{S} \delta_{S}^{\prime})$\label{line:13}\tikzmark{bottom}
    \AddNote{top}{bottom}{right}{~\hspace{-5pt}Meta-Generalizing}\\
  }
  }
  \textbf{Divide} $\bs{\mathcal{D}} \rightarrow \bs{\mathcal{D}}^{\textnormal{train}}$ and $\bs{\mathcal{D}}^{\textnormal{gen}}$\label{line:14}\\
\textsc{ttrain}$(E_0,\bs{\mathcal{D}}^{\text{train}},\bs{\mathcal{D}}^{\text{gen}})$ \label{line:15} \hfill \CommentSty{ //Warm-Up Stage}

\Whilee{not converge}{
  ~\textbf{Divide} $\bs{\mathcal{D}} \rightarrow \bs{\mathcal{D}}^{\textnormal{train}}$ and $\bs{\mathcal{D}}^{\textnormal{gen}}$ 
  \label{line:17}\\
  ~\textsc{ttrain}$(E_1,\bs{\mathcal{D}}^{\text{train}},\bs{\mathcal{D}}^{\text{gen}})$\label{line:18}

  ~\textsc{strain}$(E_2,\bs{\mathcal{D}}^{\text{train}},\bs{\mathcal{D}}^{\text{gen}})$\label{line:19}
}
\end{algorithm}

Before alternating teacher–student updates, the algorithm runs \textsc{ttrain} alone for $E_{0}$ epochs. This warm-up produces a reliable teacher and a useful shared $\Pi$, stabilizing subsequent distillation and avoiding degenerate student targets.

After warm-up, the outer loop alternates \textsc{ttrain} ($E_{1}$ epochs) and \textsc{strain} ($E_{2}$ epochs), with a fresh meta-train/meta-generalization split each time, until convergence. Reshuffling exposes updates to distribution shift and biases learning toward parameters that generalize across unseen sites and power regimes. In each cycle, the teacher is updated first, and the teacher and student steps use the same \(\bs{\mathcal D}^{\text{train}}\) and \(\bs{\mathcal D}^{\text{gen}}\). The student distills from frozen, most-recent teacher on identical data to avoid target drift; however, a capacity gap can limit how closely a small student matches a high-capacity teacher.

\subsection{DEPLOYMENT}
At deployment time, only the feature extractor ($\Pi$) and student ($\Phi$) parts of the model are retained while the teacher branch is discarded. 
Inputs are normalized ($\bar{\mathbf{H}}$) exactly as in training using \eqref{eq:normalized}, ensuring robust transmit-power adaptation.
We first generate a normalized precoding matrix $\bar{\mathbf W}$. In FDP mode, $\bar{\mathbf W}$ is obtained directly from the student model, whereas in HBF mode, we compute $\bar{\mathbf W} = \mathbf A\,\bar{\mathbf W}_{dp}$.
The final transmit precoder $\mathbf{W}$ is then obtained via \eqref{eq:precodernormalized}.

We call ``zero-shot'' the mode where the model is deployed directly to an unseen site without fine-tuning.
Alternatively, if a local CSI dataset is available, the model can be fine-tuned using a self-supervised loss. During the online fine-tuning phase, the loss is inevitably computed using noisy CSI $\hat{\mathbf{H}}$, using the loss with unlabeled local data as follows:
\begin{align}
    \mathcal{L}_{\text{site}}
    &= - {R(\bar{\mathbf{W}})}\,.
    \label{eq:site_loss}
\end{align}
To improve sample efficiency, the local CSI dataset is augmented by 
taking all possible combinations of individual user channels, 
as in \cite{karkan2024sage}. 

\section{BASELINES AND COMPLEXITY ANALYSIS} \label{Sec:Complexity}
The computational burden of mMIMO precoding algorithms represents a critical consideration alongside performance metrics. This section considers the computational energy consumption (EC) of different precoding approaches. We evaluate baseline methods against our proposed DNN-based solutions to quantify the trade-off between performance gains and implementation costs.

We consider an energy model that accounts for both the energy required by elementary operations as well as memory accesses.
This allows a fair comparison between neural networks and conventional algorithms, since the energy required to retrieve the DNN parameters from memory is accounted for.
For DNN models, we consider only the EC of inference, since the training EC becomes negligible by being amortized over a large number of uses of the model, which is reused in time and across sites.

\subsection{ENERGY MODEL FOR DNNs}
To evaluate and compare the energy of the baseline and proposed precoders, we adopt a simple yet realistic model for the EC of an application-specific hardware accelerator that accounts for both computation and memory transfers, following \cite{kasalaee2025compression, moons:2017a}. However here, for simplicity, we fix the number of bits used to represent model weights to 16. 
The computation energy $E_C$ is modeled by counting the number of \gls{MAC} operations required to compute each output activation value, including the linear operations required to compute the pre-activations as well as the bias, batch normalization, and nonlinearity computations.
It is given by
\begin{align}
    E_C=E_{\text{MAC}}(N_c+3N_a)\,,
\end{align}
where \(N_c\) denotes the total number of \gls{MAC} operations, \(N_a\) the number of activations, and the factor of \(3\) accounts for the per-activation bias, normalization, and nonlinearity computations. The energy of an elementary \gls{MAC} operation, expressed as \( E_{\text{MAC}}=0.86\) picojoules (pJ) \cite{Horowitz2014energy}. 
Based on the number of weights $N_w$ and the number of activations $N_a$ in the model, the memory-access energy is modeled for weights as
\begin{align}
    E_W=E_M N_w+E_L\,N_c/\sqrt{p}\,,     
\end{align}
and for activations as
\begin{align}
    E_A=2E_M N_a+E_L\,N_c/\sqrt{p}\,, 
\end{align}
where \(E_M\) and \(E_L\) are the energy costs of one on-chip main-memory access and one local-buffer access, respectively, and \(p=64\) 
is the effective parallelism factor. We set \(E_M=2E_L=2E_{\text{MAC}}\).
Note that for the system parameters considered in Section~\ref{Sec:Simulation}, the PaPP student model requires only about 1~MB of storage, which supports the feasibility of storing the entire model on-chip.
Combining the terms above, the total inference energy for a \gls{DNN} is given by
\begin{align}
    E_{\mathrm{DNN}}=E_C+E_W+E_A\,.
\end{align}

We adopt the approach of \cite{lyu2023downlink} as a DNN-based baseline that couples a neural model with WMMSE, but we swap the MLP for a CNN following \cite{hojatian2021unsupervised} to exploit spatial structure in \gls{CSI} and better handle larger antenna/user counts. This CNN variant (MAML-CNN) outperforms the original MAML-MLP in our experiments. Computationally, the DNN plus a final matrix inversion is heavier than \gls{ZF} yet still below WMMSE.
The number of real multiplications required by MAML-CNN is given by
\begin{align}
    N_c&^{\text{MAML-CNN}}=
18\,C_{\text{out}}N_{\sf{T}}N_{\sf{U}} + C_{\text{out}}N_{\sf{T}}N_{\sf{U}}(3N_{\sf{U}}+1) \notag\\
   &+ 8\bigg(\frac{4}{3} N_{\sf{T}}^3 + N_{\sf{T}}^2 (3N_{\sf{U}}+2) + N_{\sf{T}}(2N_{\sf{U}}+3)\bigg) \,,\\
   N_w&^{\text{MAML-CNN}}= 18\,C_{\text{out}} + (C_{\text{out}}N_{\sf{T}}N_{\sf{U}}+1)(3N_{\sf{U}}+1)\,,\\
   N_a&^{\text{MAML-CNN}}= C_{\text{out}}N_{\sf{T}}N_{\sf{U}} + 3N_{\sf{U}}+1\,,
\end{align}
where $C_{\text{in}}$ and  $C_{\text{out}}$ are the input and output channels of the CNN layer, with padding and stride of 1, and the kernel size of 3. 

The total number of real multiplications required for the PaPP student method can be expressed as
\begin{align}
N_c&^{\text{PaPP}}=
864\,N_{\sf{T}}N_{\sf{U}} + 2\,N_{\sf{T}}N_{\sf{U}}E + 4 E^2(N_{\sf{U}}+2)+4E  \notag \\
& + N_{\sf{U}}^2N_{\sf{T}}E + \underbrace{2EN_{\sf{T}}N_{\sf{U}}}_{\text{FDP}} ~\text{or}~ \underbrace{2EN_{\sf{RF}}(N_{\sf{T}}+N_{\sf{U}})}_{\text{HBF}}\,,\\
N_w&^{\text{PaPP}}= 884+N_{\sf{U}}^2N_{\sf{T}}E + 4 E^2(N_{\sf{U}}+2)+ 10 E\notag \\
& + \underbrace{2N_{\sf{T}}N_{\sf{U}}(E+1)}_{\text{FDP}} ~\text{or}~ \underbrace{2N_{\sf{RF}}(N_{\sf{T}}+N_{\sf{U}})(E+1)}_{\text{HBF}}\,,\\
N_a&^{\text{PaPP}}= 20 N_{\sf{T}}N_{\sf{U}} +N_{\sf{T}}N_{\sf{U}}^2+4EN_{\sf{U}}+10E \,.
\end{align}

\subsection{ENERGY MODELS FOR BASELINES}
The \gls{WMMSE} method, while achieving high performance, incurs substantial energy costs due to its iterative nature and repeated matrix inversions, making it considerably more energy-demanding than other methods. To ensure a fair and conservative baseline, we estimate computation energy by counting only the real multiplications and memory energy by considering only local-buffer accesses for the operands. Under these assumptions, the total number of real multiplications required for \(I\) iterations of \gls{WMMSE} is
\begin{align}
  &N_c^{\text{WM}} = \notag\\
  &4I N_{\sf{U}} \bigg(\frac{2}{3} N_{\sf{T}}^3 + N_{\sf{T}}^2 + 2N_{\sf{T}}(2N_{\sf{U}}+1) 
  + N_{\sf{U}}+\frac{14}{3}\bigg)\,,  
\end{align}
where $I$ is the total number of iterations.

The \gls{ZF} precoding method \cite{nayebi2017precoding} offers substantially lower computational complexity. This efficiency arises from its use of simpler linear algebra operations, particularly the inversion of smaller matrices when \(N_{\sf T} > N_{\sf U}\). However, despite its low computational cost, \gls{ZF} is prone to performance degradation in high-interference environments or under challenging channel conditions. The total number of real multiplications required by \gls{ZF} is
\begin{align}
    N_c^{\text{ZF}} = 8 N_{\sf{U}}^2 N_{\sf{T}} + \frac{8}{3} N_{\sf{U}}^3 \, .
\end{align}

The PE-AltMin \cite{yu2016alternating} algorithm iteratively designs \glspl{HBF} to approximate the \gls{FDP} solution by alternating between optimizing the digital baseband precoder and the analog RF precoder. 
For performance benchmarking, ZF or WMMSE are used as the FDP targets. ZF serves as a reference for a low EC choice, while WMMSE acts as the target for a high sum-rate choice. 
The number of real multiplications required to obtain the HBF precoder, including the operations necessary to compute the FDP target, is given by
\begin{align}
N_c^{\text{PE}} = N_{c}^{\text{ZF or WMMSE}} + 4TN_{\sf{RF}}\Big[\frac{4}{3} &N_{\sf{RF}}^2 +2 N_{\sf{RF}} (N_{\sf{T}} + N_{\sf{U}})\notag\\ &+ N_{\sf{T}} ( 2N_{\sf{U}} + 1) \Big] \,,
\end{align}
where $T$ corresponds to the number of iterations used for PE-AltMin.

The energy model for baseline methods can then be expressed as
\begin{equation}
    E_{\mathrm{B}}^{\text{X}} = N_c^\text{X} (E_\text{MAC} + E_L/ \sqrt{p}) \, ,
\end{equation}
where X is substituted with the desired method name.
\begin{table}[tb]
\caption{Energy consumption (EC) per use and sum rate for different methods reported for ``Ericsson'', ``Decarie'', and ``Sainte-Catherine'' sites, $N_{\sf{T}} = 64$, $N_{\sf{U}} = 4$, SNR = 10\,dB.}
\label{tab:complexity}
\resizebox{\columnwidth}{!}{%
\begin{tabular}{@{}ccccc@{}}
\toprule

\multirow{2}{*}{\textbf{Methods}} 
& \multirow{2}{*}{\textbf{EC ({$\mathbf{\mu}$}J)}} 
& \multicolumn{3}{c}{\textbf{Sum-Rates (b/s/Hz)}} \\
\cmidrule(l){3-5}
& & Ericsson & Decarie & Sainte-Catherine \\

\midrule
\midrule
\multicolumn{5}{c}{\textbf{FDP Methods}} \\
\midrule

WMMSE ($I=15$)
& 42.1 & 18.4 & 14.1 & 11.9 \\ 

MAML-CNN-FDP (zero-shot)
& 4.6 & 16.2 & 12.1 & 9.8 \\ 

PaPP-FDP (Teacher model)
& 2.5 & 18.2 & 14.1 & 11.8 \\ 

PaPP-FDP (zero-shot)
& 2.0 & 17.6 & 13.3 & 11.0 \\ 

PaPP-FDP (40 few-shot)
& 2.0 & 19.1 & 14.6 & 12.3 \\ 

PaPP-FDP (4M fine-tune)
& 2.0 & 19.9 & 14.9 & 12.6 \\ 

Zero Forcing 
& 0.01 & 11.4 & 3.9 & 2.5 \\

\midrule
\multicolumn{5}{c}{\textbf{HBF Methods}} \\
\midrule

WMMSE + PE-AltMin ($T=100$) 
& 47.5 & 14.3 & 10.7 & 9.6 \\ 

ZF + PE-AltMin ($T=100$) 
& 5.4 & 2.1 & 1.7 & 1.2 \\ 

PaPP-HBF (zero-shot)
& 2.2 & 17.2 & 13.4 & 11.9 \\ 

PaPP-HBF (40 few-shot)
& 2.2 & 17.3 & 13.5 & 12.0 \\ 

PaPP-HBF (4M fine-tune)
& 2.2 & 17.6 & 13.7 & 12.1 \\ 

\bottomrule
\end{tabular}%
}
\end{table}

Table~\ref{tab:complexity} summarizes the EC and sum-rate performance across different sites. For the FDP methods, quantitatively, the proposed PaPP method reduces EC by approximately $20\times$ compared to WMMSE, while achieving comparable sum-rates. Furthermore, PaPP is twice as energy-efficient as the MAML-CNN baseline. For the fine-tuned and few-shot variants, we report only the inference energy, as the one-time computational cost of adaptation is amortized over a large number of subsequent transmission intervals and can thus be neglected. Although ZF has negligible computational cost ($0.01\,\mu$J), its performance degrades severely in complex environments (e.g., Sainte-Catherine); in contrast, PaPP incurs a modest energy overhead ($2.0\,\mu$J) to deliver robust interference mitigation and high spectral efficiency. For the HBF methods, while ZF + PE-AltMin offers a low EC baseline, it is still significantly more energy-intensive than the proposed PaPP-HBF and suffers from poor sum-rate performance. Conversely, WMMSE + PE-AltMin targets a high sum-rate but ultimately yields a lower sum-rate than our PaPP-HBF method. Consequently, PaPP-HBF dominates both conventional baselines, achieving superior spectral efficiency across all sites while reducing EC by over $21\times$, compared to the WMMSE + PE-AltMin benchmark.

\section{NUMERICAL RESULTS} \label{Sec:Simulation}
We evaluate PaPP on a custom ray-tracing dataset across held-out deployment sites and a wide \gls{SNR} range, reporting spectral-efficiency results under both \gls{FDP} and \gls{HBF} settings. We compare zero-shot performance, self-supervised few-shot adaptation, and full fine-tuning against \gls{ZF}, \gls{WMMSE}, PE-AltMin, MAML-CNN, DeepAll-CNN, and a site/\gls{SNR}-specific student trained from scratch. The evaluation isolates site/domain shift and \gls{SNR} robustness, and quantifies throughput–energy trade-offs by emphasizing inversion-free, non-iterative inference and compact models.
DeepAll-CNN is based on the method of \cite{hojatian2021unsupervised}, which uses convolutional layers for feature extraction followed by direct precoder design. The DeepAll-CNN is trained by merging all MLDG domains into a single training set and using the standard self-supervised procedure.

All \glspl{BS} use an $8 \times 8$ uniform planar array ($N_{\sf{T}} = 64$) 
with $\lambda/2$ spacing at a 2 GHz carrier frequency. User positions are 
sampled around the BS in polar coordinates (radii 50-350 m with 
$\ang{10}$ azimuth steps), and each dataset entry aggregates 
$N_{\sf{U}}=4$ user locations.
Table~\ref{tab:hyper} lists the PaPP backbone training hyperparameters, which were optimized using a validation set.

\vspace{-10pt}
\subsection{RAY-TRACING DATASET}
In this work, we introduce a custom dataset designed to model channel characteristics in a \gls{mMIMO} system and evaluate the performance of the PaPP, focusing on its adaptability and efficiency in simulated real-world wireless environments. The dataset is generated using the MATLAB Ray-Tracing toolbox, simulating realistic propagation conditions with both \gls{LOS} and \gls{NLOS} components. 
Base stations are deployed in several locations in the greater Montreal area, utilizing OpenStreetMap (OSM) \cite{OpenStreetMap} data to incorporate real-world structures and materials. The propagation environment is configured to account for up to 10 reflections with up to 1 diffraction. This setup ensures the dataset accurately captures realistic multipath characteristics influenced by building structures, terrain, and other environmental factors. The training dataset encompasses diverse locations, including areas such as ``Université de Montréal'', ``Parc'', ``Rachel'', ``Cathcart'', ``Old Port'', ``Sherbrooke'', and ``Okapark'', ensuring the model generalizes effectively across different environments. In this paper, we exclude three deployment datasets from the training and use them to test the generalization of the methods: ``Ericsson'' features an industrial environment with 75\% \gls{LOS} users, ``Decarie'' is a residential area with a balanced \gls{LOS}/\gls{NLOS} mix, and ``Sainte-Catherine'', a downtown area, has 75\% \gls{NLOS} users, providing a rigorous test for model adaptability.

\begin{table}[t]
    \caption{Hyperparameter settings for PaPP model training.}
    \centering
    \resizebox{\columnwidth}{!}{
    \begin{tabular}{cc}
        \toprule
        \textbf{Hyperparameter} & \textbf{Value}  \\
        \midrule
        Self-supervision weight ($\lambda_0,\,\lambda_1$) & $0.01,\,0.1$\\
        Teacher model LRs ($\alpha_{T}, \beta_{T},\epsilon_{T}$) & $10^{-1}, 10^{-2}, 10^{-2}$\\
        Feature model LRs ($\alpha_{F},\beta_{F},\epsilon_{F}$) & $10^{-1}, 10^{-2}, 10^{-2}$\\
        Student model LRs ($\alpha_{S},\beta_{S},\epsilon_{S}$) & $10^{-2}, 10^{-3}, 10^{-3}$\\
        Number of sites in each set $|\bs{\mathcal{D}}|, |\bs{\mathcal{D}}^{\text{train}}|, |\bs{\mathcal{D}}^{\text{gen}}|$ & 56, 40, 16\\
        Dataset size of each site $|\hat{\mathcal{D}}|$ & 1\,M\\
        Batch size & 1000\\
        \bottomrule
    \end{tabular}}
    \label{tab:hyper}
\end{table}

\begin{figure*}[t]
    \centering
    \subfigure[Sum rate achieved at an average SNR of 25\,dB, FDP.]{\includegraphics[width=.9\columnwidth]{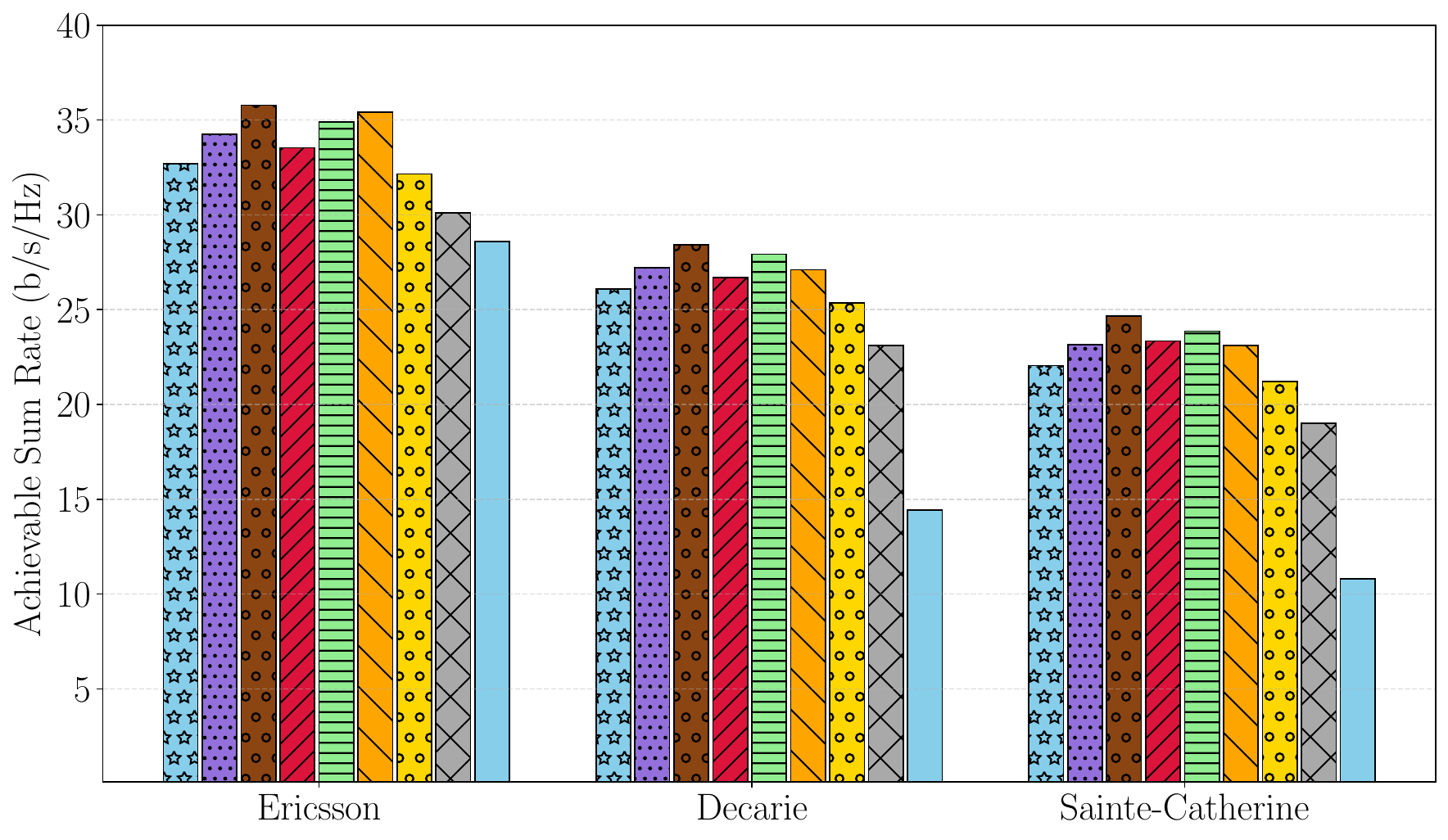}
    \label{fig:FDP_bar:25}}
    \hfill
    \subfigure[Sum rate achieved at an average SNR of 10\,dB, FDP.]{\includegraphics[width=.9\columnwidth]{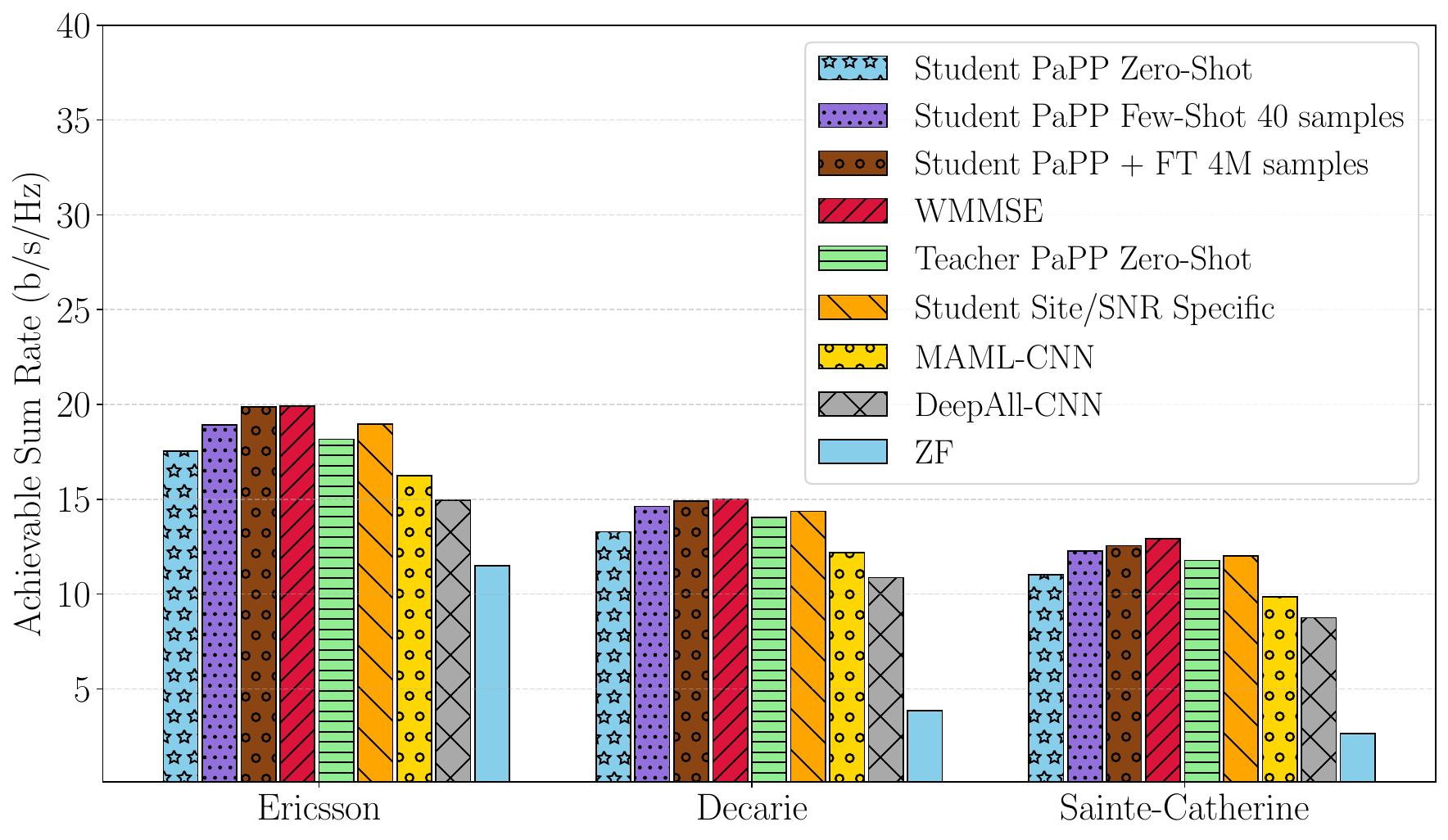}
    \label{fig:FDP_bar:10}}
    \subfigure[Sum rate achieved at an average SNR of 25\,dB, HBF.]{\includegraphics[width=.9\columnwidth]{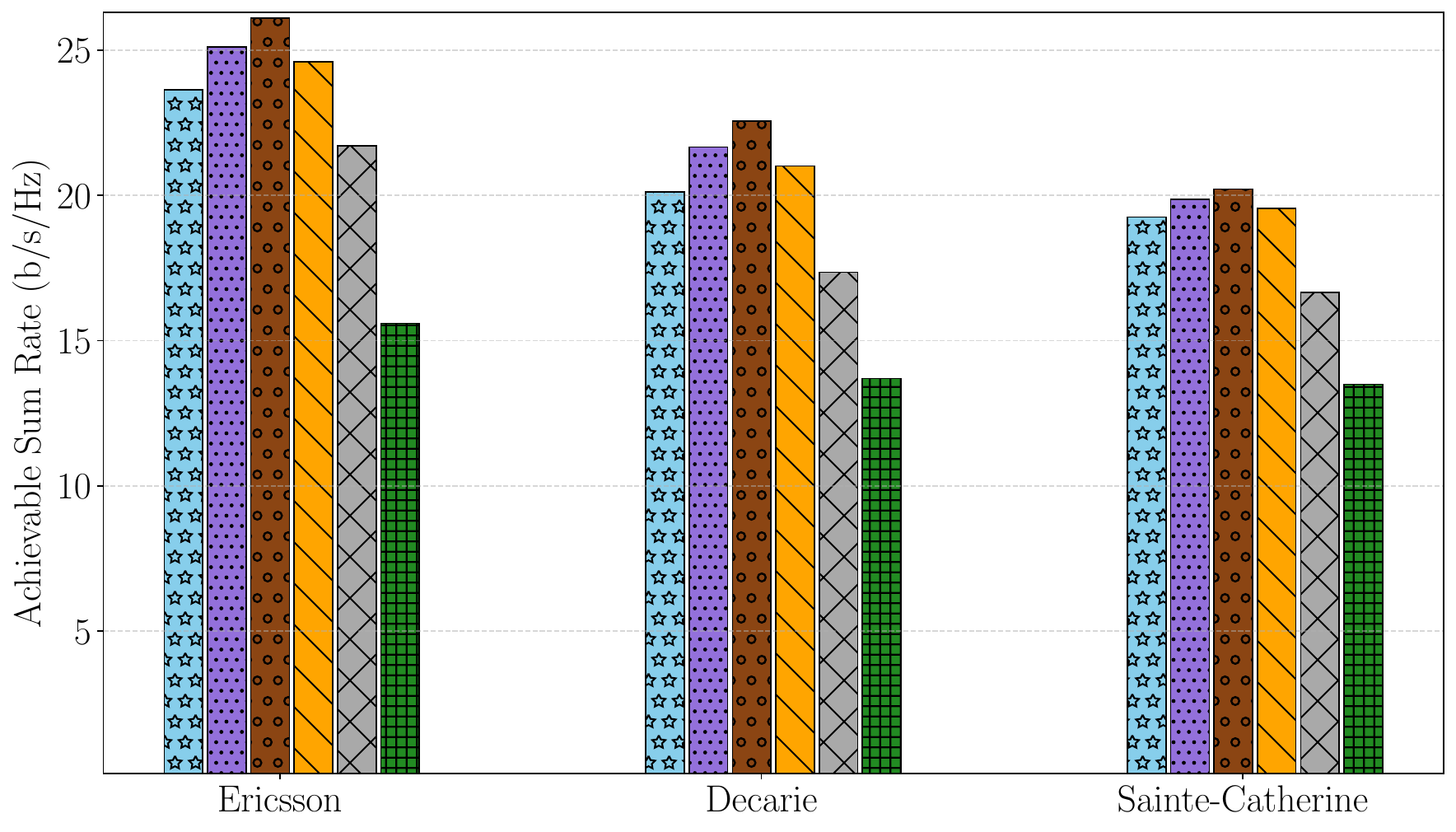}
    \label{fig:HBF_bar:25}}
    \hfill
    \subfigure[Sum rate achieved at an average SNR of 10\,dB, HBF.]{\includegraphics[width=.9\columnwidth]{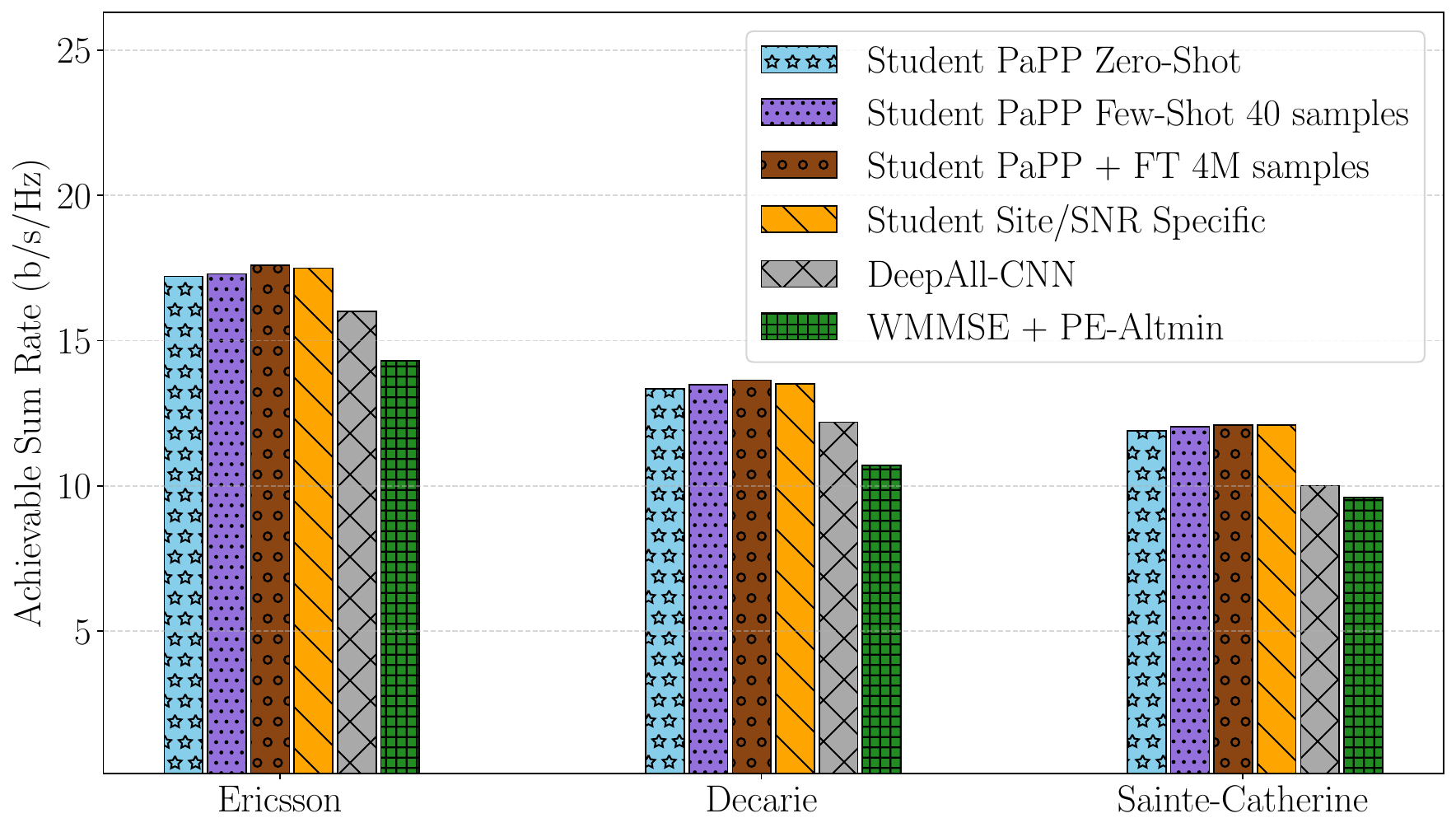}
    \label{fig:HBF_bar:10}}
    \caption{Achievable sum rates for \gls{FDP} and \gls{HBF} methods under different SNR values at three Montreal sites: ``Ericsson'', ``Decarie'', and ``Sainte-Catherine''. All methods report zero-shot performance, except ``PaPP + FT'' and ``PaPP Few-Shot'', which are after 20 epochs of fine-tuning.}
    \label{fig:FDP_bar}
\end{figure*}

\begin{figure*}[t]
    \centering
    \subfigure[Ericsson site, FDP.]{\includegraphics[width=.85\columnwidth]{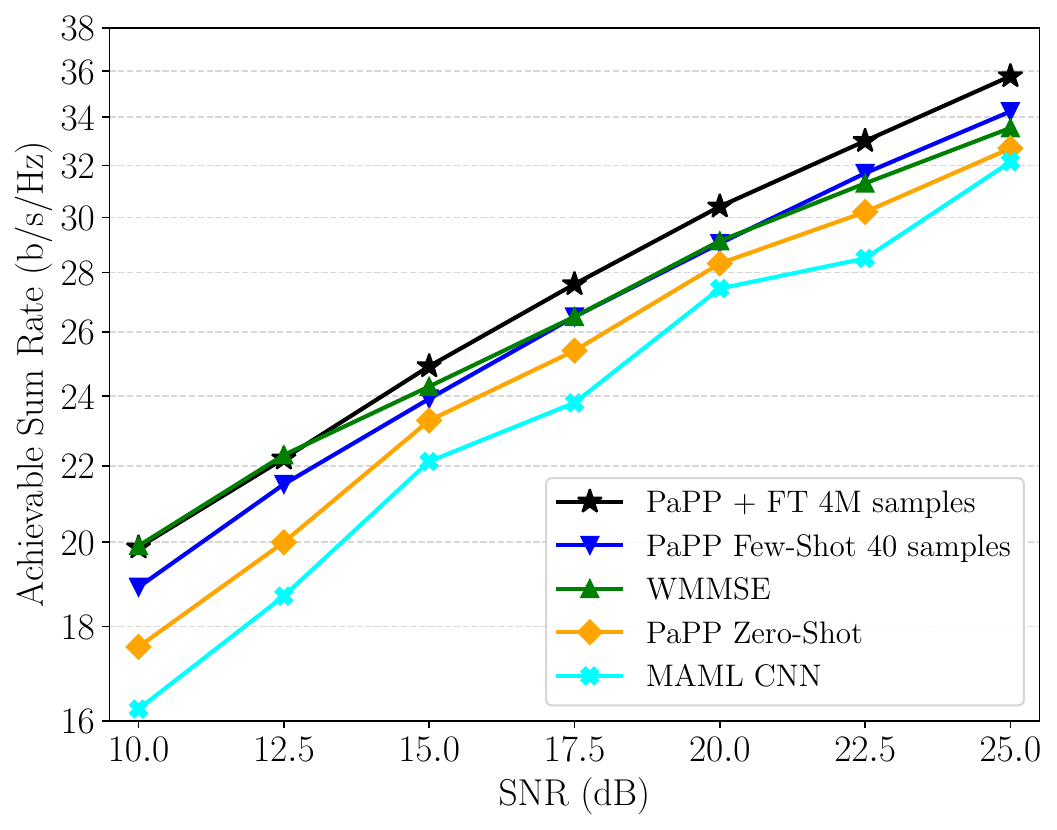}
    \label{fig:FDP:E}}
    \hfill
    \subfigure[Sainte-Catherine site, FDP.]{\includegraphics[width=.85\columnwidth]{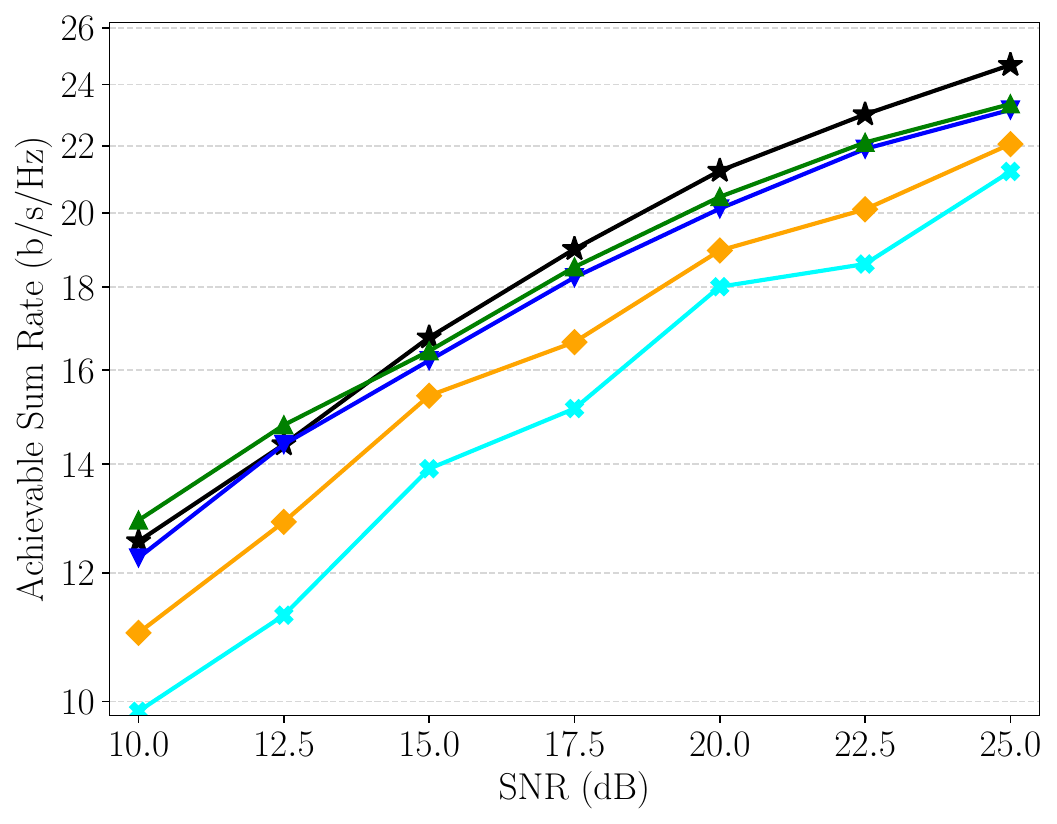}
    \label{fig:FDP:S}}
    \subfigure[Ericsson site, HBF.]{\includegraphics[width=.85\columnwidth]{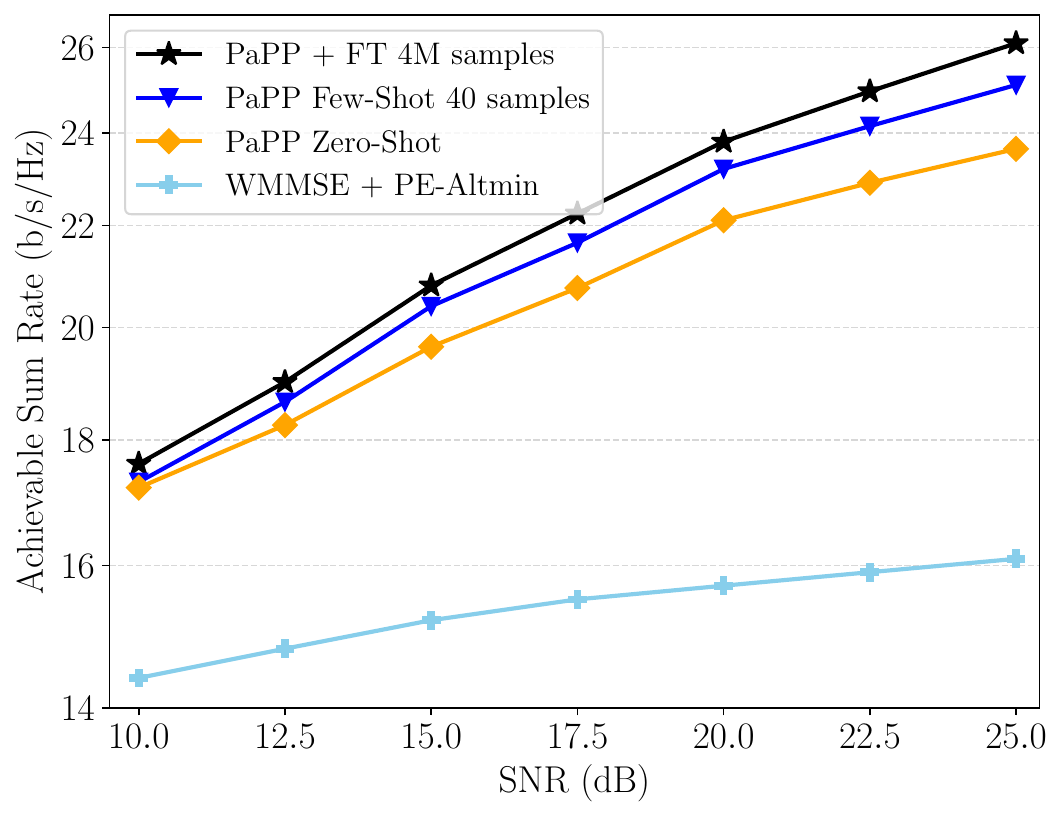}
     \label{fig:HBF:E}}
     \hfill
    \subfigure[Sainte-Catherine site, HBF.]{\includegraphics[width=.85\columnwidth]{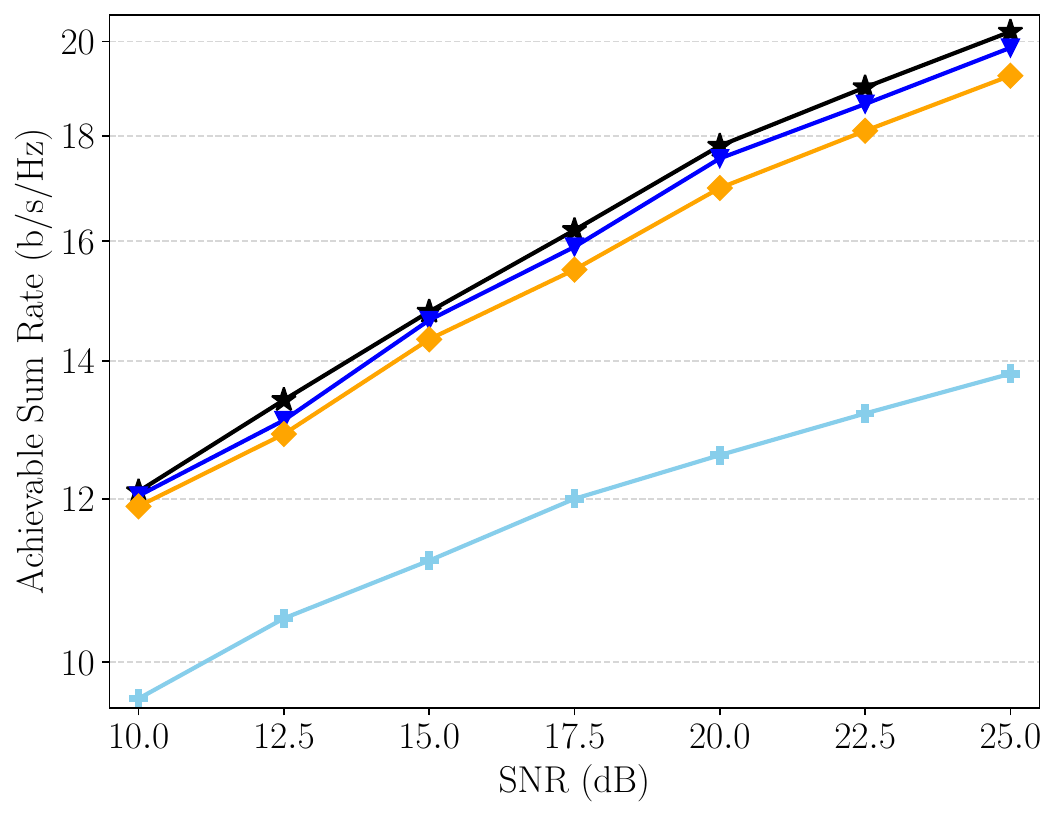}
     \label{fig:HBF:S}}
    \caption{\justifying Comparing achievable sum rates for \gls{FDP} and \gls{HBF} methods under different unseen deployment sites and SNR conditions. All methods report zero-shot performance, except ``PaPP + FT'' and ``PaPP Few-Shot'', which are after 20 epochs of fine-tuning.}
    \label{fig:FDP}
\end{figure*}

\subsection{RESULTS}
We empirically observe that \gls{DL}-based precoders trained on a single site are highly sensitive to \gls{BS} location shifts, suffering substantial degradation when deployed zero-shot at new sites. For example, using the unsupervised \gls{DL} method of \cite{hojatian2021unsupervised} as a representative baseline, training on a downtown Montréal site (Old Port) and deploying at another nearby downtown site (Sainte-Catherine) reduces zero-shot performance by \(\mathbf{53}\%\) relative to the in-domain baseline. This illustrates the severity of site/domain shift even within ostensibly similar urban environments.

We analyze the performance of the PaPP model in comparison with \gls{ZF} \cite{nayebi2017precoding}, \gls{WMMSE} \cite{shi2011iteratively} (with stopping criteria of $10^{-3}$ and maximum number of iteration of 100), the MAML-CNN, and the student site/SNR specific model, which is trained from scratch on the deployment site. The results are shown in Figures~\ref{fig:FDP_bar}, \ref{fig:FDP}, and \ref{fig:beta}. Unless stated otherwise, the results assume $\beta=0$. 

\subsubsection{Results for different sites}
In Fig.~\ref{fig:FDP_bar:25} at \(\mathrm{SNR}=25\,\mathrm{dB}\), the proposed PaPP model clearly dominates both classical and learning-based baselines across all three sites. 
The behavior of the other learning-based baselines underlines that this gain is not automatic for DNN precoders. MAML-CNN and DeepAll-CNN consistently underperform WMMSE, typically losing 4--9\% and 10--19\% sum-rate, respectively, depending on the site. In contrast, PaPP + FT improves the average sum-rate by 13\% relative to MAML-CNN and more than 20\% relative to DeepAll-CNN. Simply aggregating all different domains into a single training dataset (DeepAll) or relying on generic gradient-based meta-learning (MAML) is therefore insufficient to cope with the strong cross-site variations. 
The different PaPP operating modes also illustrate a clear accuracy–data trade-off. The PaPP teacher in zero-shot mode already outperforms WMMSE at all three sites, with 3--5\% higher sum-rate on average, and is competitive with or better than the “Student Site/SNR Specific” model trained from scratch on the target site. The student in zero-shot mode starts a few percent below WMMSE, but even a very small amount of local data is enough to close much of this gap: with only 40 samples per site (“Student PaPP Few-Shot”), the student recovers roughly half of the improvement between zero-shot and fully fine-tuned performance and reaches or slightly exceeds WMMSE on Ericsson and Decarie, while remaining within 1\% of WMMSE on Sainte-Catherine. When large local datasets are available, fine-tuning the PaPP student with 4M samples yields up to 7\% higher sum-rate than training the same student architecture in a purely site/SNR-specific manner, indicating that the multi-site backbone does not just give a good initialization but leads to better local optima after adaptation.

In Fig.~\ref{fig:FDP_bar:10} at \(\mathrm{SNR}=10\,\mathrm{dB}\), in the noise-dominated 10\,dB regime, classical \gls{WMMSE} retains an advantage. Yet the PaPP backbone transfers robustly across sites and clearly outperforms all other learning-based baselines. 
Across sites, the fine-tuned student (“PaPP + FT 4M”) tracks WMMSE almost exactly, it is within 0.04 b/s/Hz on Ericsson, 0.12 on Decarie, and 0.38 on Sainte-Catherine, leading to an average sum-rate only about 1\% below WMMSE. This shows that the compressed student can essentially recover WMMSE-level performance even at low SNR, where optimization-based methods are strongest.
The few-shot student with only 40 local samples already reaches 95--99\% of the WMMSE performance at all sites and is consistently better than the site/SNR-specific student trained from scratch. The PaPP teacher and student in zero-shot mode form a smooth performance ladder below that, but still clearly above the generic MAML-CNN and DeepAll-CNN baselines, which lag WMMSE by more than 3 b/s/Hz on average. ZF collapses at 10 dB, especially in Decarie and Sainte-Catherine, confirming that linear precoding is highly suboptimal in these interference-limited regimes.

In Fig.~\ref{fig:HBF_bar:25} at \(\mathrm{SNR}=25\,\mathrm{dB}\) with hybrid beamforming, the PaPP student dominates all HBF baselines at each site. The fully fine-tuned model achieves the highest sum rate on Ericsson, Decarie, and Sainte-Catherine, beating both the classical PE-Altmin algorithm and the DeepAll-CNN baseline by a comfortable margin. Even the simpler “Student Site/SNR Specific” model consistently outperforms PE-Altmin, showing that data-driven HBF can clearly surpass the conventional alternating-minimization design when enough training data is available. 
The different PaPP regimes show strong generalization and data efficiency. In zero-shot mode, the PaPP student already surpasses both DeepAll-CNN and PE-Altmin at all three sites, which means the shared backbone learned across sites transfers well to new HBF deployments without any local retraining. With only 40 labeled samples per site, the few-shot student moves close to the fully fine-tuned performance and stays strictly above both site-specific training and PE-Altmin everywhere.

In Fig.~\ref{fig:HBF_bar:10} at \(\mathrm{SNR}=10\,\mathrm{dB}\) with hybrid beamforming, PaPP stays on top, but the margins compress. The fully fine-tuned student achieves the highest sum rate at all three sites. Zero-shot PaPP is already better than PE-Altmin everywhere, and the few-shot model with only 40 samples adds a small but consistent gain on top of that. This shows that the shared backbone generalizes well even in the more noise-limited 10 dB regime, and only modest local adaptation is needed to reach near-optimal performance.
The gap between PaPP and the site/SNR-specific student is also small at 10 dB. In contrast, DeepAll-CNN is clearly weaker than all PaPP variants across sites.

\subsubsection{Results with varying transmit power}
\begin{figure*}[t]
    \centering
    \subfigure[\justifying Comparison with baseline methods.]{\includegraphics[width=.85\columnwidth]{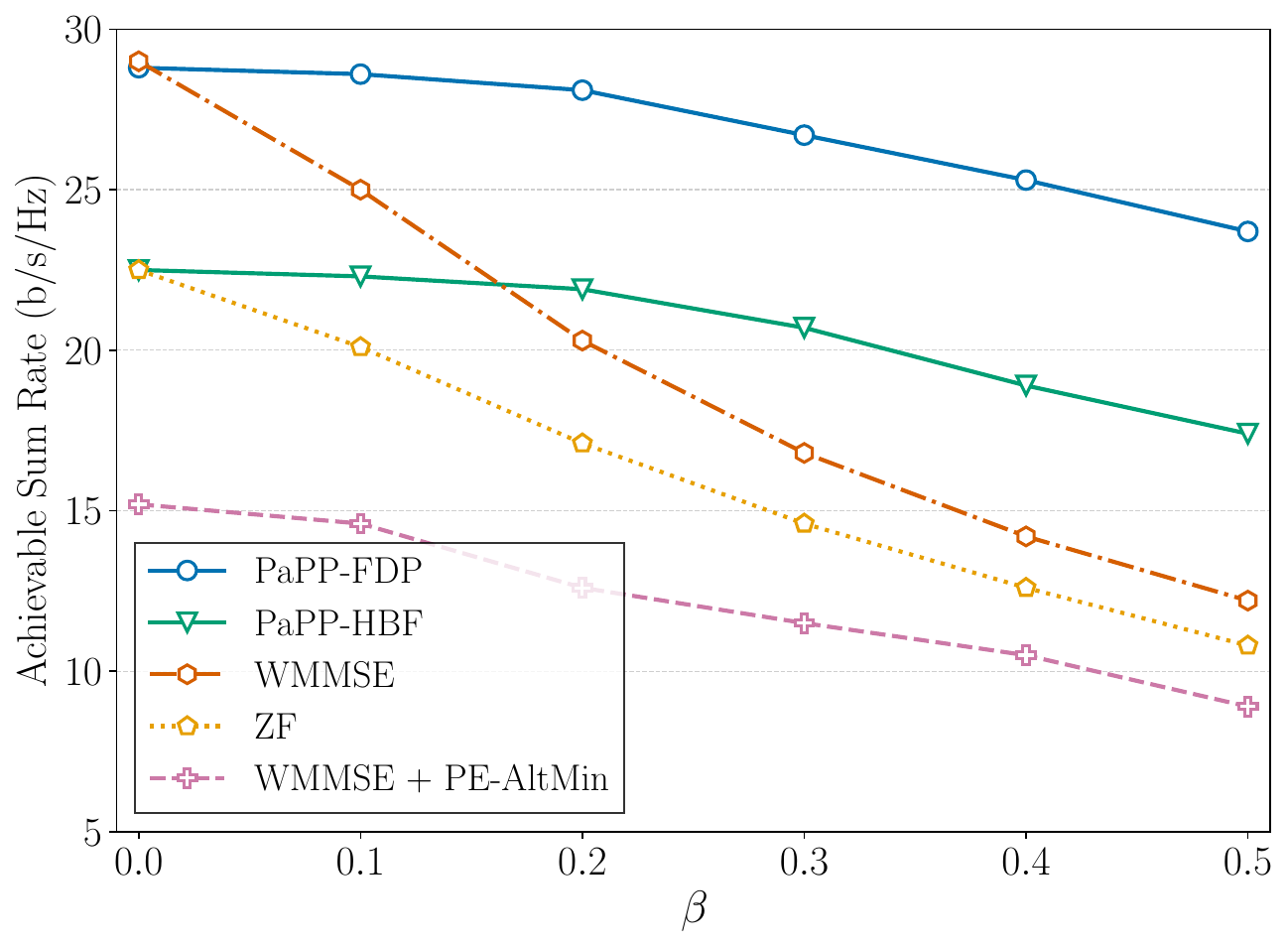}\label{fig:beta:ZS}}
    \hfill
    \subfigure[\justifying Impact of noisy local CSI on fine-tuning.]{\includegraphics[width=.85\columnwidth]{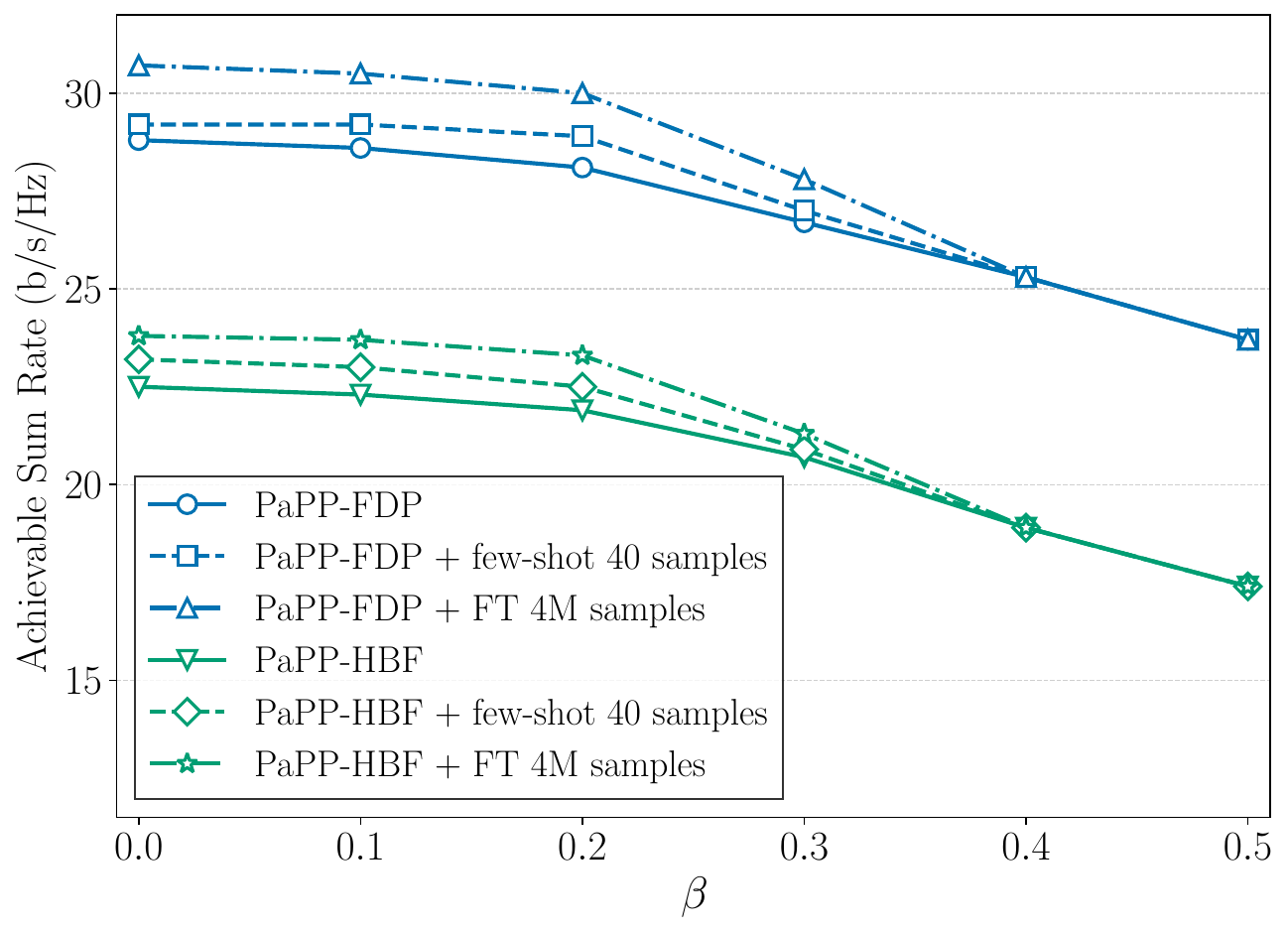}\label{fig:beta:FT}}
    \caption{\justifying Impact of the channel estimation noise factor $\beta$ on the performance of the proposed PaPP models during deployment and fine-tuning for the ``Ericsson'' site at an average SNR of 20\,dB.}
    \label{fig:beta}
\end{figure*}

In Fig.~\ref{fig:FDP:E} at the Ericsson site, the full fine-tuning student PaPP achieves high spectral efficiency, effectively matching WMMSE at 10~dB and consistently outperforming it from 15~dB upward by up to 7\%. The few-shot student with only 40 samples remains very competitive, nearly matching WMMSE at 20~dB and slightly surpassing it at 25~dB, while the zero-shot student tracks WMMSE closely across SNRs, staying within 12\% at 10~dB and within 3--4\% for 15--25~dB. Over the same range, the zero-shot student consistently outperforms the MAML-CNN baseline by 2--8\%. This demonstrates that the shared PaPP backbone yields a strong zero-shot precoder that is already competitive with iterative optimization methods and significantly outperforms conventional meta-learning, even before any local adaptation. In Fig.~\ref{fig:FDP:S} at the Sainte-Catherine site, on this more challenging downtown site, the fully fine-tuned student PaPP demonstrates robust adaptation. It starts slightly below WMMSE at $\mathrm{SNR}=10\,\mathrm{dB}$ but surpasses it by $\mathrm{SNR}=15\,\mathrm{dB}$, maintaining a performance gain of approximately 2--6\% at higher SNRs. The few-shot student, with only 40 samples, remains within a few percent of the performance of WMMSE.

For HBF, in Fig.~\ref{fig:HBF:E} on the Ericsson site, the proposed PaPP-based methods consistently outperform the conventional PE-Altmin baseline across all SNR values. In zero-shot mode, PaPP achieves improvements of 20--47\% over PE-Altmin, with gains widening substantially as SNR increases. With only 40 fine-tuning samples, the few-shot variant increases these gains to 20--56\%. The fully fine-tuned PaPP model yields the largest improvements, outperforming the baseline by approximately 22--62\%. 
In Fig.~\ref{fig:HBF:S} on the dense downtown Sainte-Catherine site, all PaPP variants continue to consistently outperform the PE-Altmin baseline across the SNR range. In zero-shot mode, PaPP achieves 24--40\% higher sum-rate than PE-Altmin. With only 40 samples, the few-shot variant raises these improvements to 25--44\%, while the fully fine-tuned PaPP model provides the largest gains of roughly 26--47\% relative to PE-Altmin. In terms of data efficiency, the meta-trained backbone generalizes well, the zero-shot model attains 95--98\% of the fully fine-tuned PaPP performance, and adaptation with only 40 local samples recovers 98--99.5\% of full fine-tuning. 

The evaluation of models at intermediate SNR values (12.5, 17.5, and 22.5 dB), which were unseen during the backbone training of either PaPP or MAML-CNN, reveals smooth performance scaling with respect to SNR. While MAML-CNN exhibits a performance drop at these unseen SNRs, the PaPP models maintain a strictly monotone increase in both the easier (Ericsson) and more challenging (Sainte-Catherine) environments. This performance confirms that the backbone’s transmit power-aware normalization and \gls{MLDG} yield a stable mapping capable of generalizing across the full operating SNR range.

\subsubsection{Results for Robustness to Noisy CSI}
We focus on two practically distinct use cases. First, robustness when the test-time CSI noise is unknown and may differ from training, and second, site-specific fine-tuning when only noisy CSI is available at the deployment site. 

To begin with, we train the backbone model using
$\beta \in \mathcal{B}_\textsc{bb} = \{0,\,0.2,\,0.5\}$.
In Fig.~\ref{fig:beta:ZS}, on the Ericsson site at \(\mathrm{SNR}=20\,\mathrm{dB}\), 
while the sum-rate of Near-optimal FDP and WMMSE drops by more than 50\% as $\beta$ increases from 0 to 0.5, PaPP-FDP maintains a stable performance, decreasing by only 18\%. At $\beta=0.2$, PaPP-FDP improves the sum-rate by 22\% relative to Near-optimal FDP and by 38\% relative to WMMSE. As the estimation error becomes more severe, these gains widen significantly: at $\beta=0.5$, PaPP-FDP achieves a 59\% higher sum-rate than Near-optimal FDP and a 94\% improvement over WMMSE. A similar trend is observed for the hybrid architecture, PaPP-HBF consistently outperforms PE-AltMin, with gains ranging from 7\% under perfect CSI conditions to 27\% at $\beta=0.5$.

To begin with, we train the backbone model using $\beta \in \mathcal{B}_\textsc{bb} = \{0,\,0.2,\,0.5\}$. In Fig.~\ref{fig:beta:ZS}, on the Ericsson site at $\mathrm{SNR}=20\,\mathrm{dB}$, while the sum-rate of WMMSE drops by roughly 58\% as $\beta$ increases from 0 to 0.5, PaPP-FDP maintains a stable performance, decreasing by only 18\%. At $\beta=0.2$, PaPP-FDP improves the sum-rate by 38\% relative to WMMSE and by 64\% relative to ZF. As the estimation error becomes more severe, these gains widen significantly: at $\beta=0.5$, PaPP-FDP achieves a 94\% higher sum-rate than WMMSE and a 119\% improvement over ZF. A similar trend is observed for the hybrid architecture, where PaPP-HBF consistently outperforms PE-AltMin, with gains ranging from 48\% under perfect CSI conditions to 96\% at $\beta=0.5$.

Furthermore, in Fig.~\ref{fig:beta:FT}, when we deploy to the deployment site, only noisy CSI with level $\beta_\textsc{ft}$ is available for fine-tuning. For low-to-moderate noise levels ($\beta_\textsc{ft} \leq 0.3$), few-shot adaptation with just 40 noisy samples yields a 1--3\% improvement over the zero-shot baseline. Full fine-tuning with $4$~M samples further extends these gains to approximately 5--7\%, with similar trends observed for the PaPP-HBF architecture. This confirms that the backbone model can effectively extract useful features even from imperfect local CSI. However, as expected, the benefit of adaptation diminishes as the estimation error becomes severe. For $\beta_\textsc{ft} \geq 0.4$, the performance of both few-shot and fully fine-tuned models converges to that of the zero-shot backbone. Specifically, at $\beta_\textsc{ft}=0.5$, fine-tuning provides no sum-rate gain, indicating that the training signal provided by excessively corrupted CSI is dominated by noise, preventing further refinement of the precoders beyond the robust baseline established by the backbone.

\vspace{-10pt}
\section{CONCLUSION} \label{Sec:conclusion}
This work proposed a computation-efficient deep learning framework that leverages a meta-trained teacher-student architecture and power-aware normalization to generalize across diverse deployment scenarios. The approach outperforms both conventional iterative algorithms and existing deep learning benchmarks in unseen environments and maintains near-optimal spectral efficiency across a wide range of transmit power levels. Furthermore, the model demonstrates excellent robustness to channel estimation errors, yielding sum-rate gains of up to 94\% over conventional methods under severe channel estimation errors. These performance advantages are achieved while reducing compute energy by at least 21\(\times\) compared to conventional baselines, and requiring as few as 40 local samples for near-optimal adaptation.
\bibliographystyle{IEEEtran}
\bibliography{main}

\end{document}